%% file: main.tex
\documentclass[letterpaper]{article}
\usepackage[preprint]{acl}

\usepackage{times}
\usepackage{latexsym}
\usepackage[T1]{fontenc}
\usepackage[utf8]{inputenc}
\usepackage{microtype}
\usepackage{inconsolata}
\usepackage{graphicx}

\input{aaai_preamble}

\title{\sys{}: Multi-Agent Systems for\\ Structured Benchmark Creation}

\author{
  \textbf{Natasha Butt$^{1}$\thanks{Work done while at Microsoft Research. Correspondence to n.e.butt@uva.nl and 
vidhishab@microsoft.com.}, Varun Chandrasekaran$^{2,3}$, Neel Joshi$^{2}$,}\\
  \textbf{Besmira Nushi$^{2}$, Vidhisha Balachandran$^{2}$} \\
  $^1$University of Amsterdam $^2$Microsoft Research, $^3$UIUC \\
}

\begin{document}
\setcounter{secnumdepth}{3}

\maketitle

\begin{abstract}
\input{sections/0_abstract}
\end{abstract}

\input{sections/1_introduction}
\input{sections/2_related_work}
\input{sections/3_components}
\input{sections/4_benchmark_generation}
\input{sections/5_quality}

\input{sections/6_trial}

\input{sections/9_conclusion}

\section*{Acknowledgements}
We sincerely thank Siddharth Joshi, Arindam Mitra, Mazda Moayeri, Eduardo Salinas, Alessandro Stolfo, Vibhav Vineet, and Safoora Yousefi for their invaluable feedback and insightful discussions during this project.

\newpage
\bibliography{main}

\newpage
\appendix
\onecolumn
\section*{Appendix}

\input{sections/appendix/_10_appendix}

\end{document}

%% file: aaai_preamble.tex
\usepackage{times}
\usepackage{helvet}
\usepackage{courier}
\usepackage{natbib}
\usepackage{caption}
\usepackage{algorithm}
\usepackage{algorithmic}
\usepackage{newfloat}
\usepackage{listings}

\usepackage{times}
\usepackage{latexsym}
\usepackage[T1]{fontenc}

\usepackage{fontawesome5} 
\usepackage{array}        
\usepackage{makecell}     
\usepackage{xcolor}       
\input{math_commands.tex}

\usepackage{tabularx}
\usepackage{booktabs}
\usepackage{url}
\usepackage{graphicx} 
\usepackage{xcolor}
\usepackage{xspace}
\usepackage{subcaption}
\usepackage{graphicx}
\usepackage{float}
\usepackage{inconsolata}

\usepackage{lineno}

\definecolor{darkblue}{rgb}{0, 0, 0.5}
\usepackage{tcolorbox}

\newcounter{myboxcounter}
\renewcommand{\themyboxcounter}{\arabic{myboxcounter}}

\newenvironment{mybox}[1][]{%
  \refstepcounter{myboxcounter}%
  \begin{tcolorbox}[title=Example \themyboxcounter: #1]%
}{%
  \end{tcolorbox}%
}

\DeclareCaptionStyle{ruled}{labelfont=normalfont,labelsep=colon,strut=off}
\floatstyle{ruled}
\newfloat{listing}{tb}{lst}
\floatname{listing}{Listing}
\newcommand{\sys}[1]{\textsc{BenchAgents}\xspace}
\newcommand{\taskone}[1]{BA-Calendar\xspace}
\newcommand{\tasktwo}[1]{BA-Text\xspace}
\newcommand{\taskthree}[1]{BA-Causal\xspace}
\newcommand{\vagent}[1]{V-Agent\xspace}
\newcommand{\gagent}[1]{G-Agent\xspace}
\newcommand{\pagent}[1]{P-Agent\xspace}
\newcommand{\eagent}[1]{E-Agent\xspace}
\definecolor{darkgreen}{rgb}{0.0, 0.5, 0.0}

\newif\iffeedback
\feedbacktrue

\iffeedback
\newcommand{\varun}[1]{\textcolor{red}{Varun: #1}}
\newcommand{\vidhisha}[1]{\textcolor{teal}{Vidhisha: #1}}
\newcommand{\tash}[1]{\textcolor{blue}{Tash: #1}}
\newcommand{\besa}[1]{\textcolor{purple}{Besa: #1}}
\newcommand{\todo}[1]{\textcolor{orange}{TODO: #1}}
\newcommand{\VV}[1]{\textcolor{magenta}{VV: #1}}
\else
\newcommand{\varun}[1]{}
\newcommand{\vidhisha}[1]{}
\newcommand{\tash}[1]{}
\newcommand{\besa}[1]{}
\newcommand{\todo}[1]{}
\newcommand{\VV}[1]{}
\fi

\newcounter{myctr}

\newenvironment{myitemize}{\begin{list}{$\bullet$}
{\setlength{\topsep}{1mm}\setlength{\itemsep}{0.25mm}
\setlength{\parsep}{0.1mm}
\setlength{\itemindent}{0mm}\setlength{\partopsep}{0mm}
\setlength{\labelwidth}{15mm}
\setlength{\leftmargin}{4mm}}}{\end{list}}

%% file: math_commands.tex
\usepackage{amsmath,amsfonts,bm}

\def\eqref#1{equation~\ref{#1}}

\def\1{\bm{1}}

\DeclareMathAlphabet{\mathsfit}{\encodingdefault}{\sfdefault}{m}{sl}
\SetMathAlphabet{\mathsfit}{bold}{\encodingdefault}{\sfdefault}{bx}{n}



%% file: sections/0_abstract.tex
Evaluation insights are limited by the availability of high-quality benchmarks. As models evolve, there is a need to create benchmarks that can measure progress on new and complex generative capabilities. However, manually creating new benchmarks is slow and expensive, restricting comprehensive evaluations for any capability. We introduce \sys{}, a multi-agent framework that methodically leverages large language models (LLMs) to automate evaluation benchmark creation while inherently ensuring data and (evaluation) metric quality. \sys{} decomposes the benchmark creation process into planning, generation, verification, and evaluation, each of which is orchestrated via LLM agents. These agents interact with each other and utilize feedback from benchmark developers to improve and flexibly control data diversity and quality. We use \sys{} to create benchmarks to evaluate capabilities related to planning, constraint satisfaction, and causal reasoning spanning both language and vision modalities. We then use these benchmarks to study state-of-the-art models and extract new insights into common failure modes and model differences.

%% file: sections/1_introduction.tex
\section{Introduction}
\label{sec:intro}

\input{figures/tex/comparison}
AI advancements are progressing rapidly, with new models frequently showing enhanced capabilities. Evaluation datasets are essential for testing these claims, but quickly saturate~\citep{balachandran2024eureka}, and/or be contaminated~\citep{zhang2024languagemodeldevelopersreport}. In the absence of benchmarks, new capabilities are often demonstrated with anecdotal, qualitative examples or small, non-comprehensive test sets. This highlights the need for scalable, dynamic benchmarking methods. 
Traditionally, benchmark creation involved designing data requirements and recruiting human annotators to provide instances. Whilst often ensuring quality, this process is costly, time-consuming, and difficult to scale. Previous work proposed methods for the generation of synthetic evaluation data. However, they are not generalizable: requiring customized prompt templates~\citep{wang2024benchmarkselfevolvingmultiagentframework, xia2024leaderboardrankingcoding, yuan2024sevalautomaticadaptivetest}, or using programmatic workflows for select domains~\citep{zhu2024dyvaldynamicevaluationlarge, zhang2024task} often grounded in existing benchmarks. 

We propose \sys{} a {\em multi-agent framework} for automated, high-quality, and diverse benchmark creation and evaluation. \sys{} automates the benchmark creation process using four components (see Fig.~\ref{fig:framework}).
The Planning Agent creates a structured benchmark design/plan, and communicates the plan with the other agents. The plan contains (i) \emph{parameters and constraints} that influence dataset diversity, (ii) definitions for \emph{verification checks} to validate quality of generated data, and (iii) comprehensive \emph{metrics} for model evaluation and dis-aggregations along important dimensions. The Data Generation Agent implements the plan and produces code for scalable and controllable synthetic data; grounding the generation on a structured plan enables precise control of data diversity. It can be augmented to use external tools or libraries to further enrich the generated data. Next, the Verification Agent implements the fine-grained data quality checks to ensure quality control for the generated examples. Finally, the Evaluation Agent implements the proposed evaluation metrics to comprehensively evaluate the performance of the target model. \sys{} by design allows for additional developer feedback at each stage of the process to ensure transparency, control and quality in the produced benchmarks. 

\sys{} is simple in design and extensible to accommodate diverse tasks and modalities.
We demonstrate its utility, by generating benchmarks for three complex tasks across language and vision: calendar scheduling (\taskone{}), long-form text generation following constraints (\tasktwo{}) and visual causal reasoning (\taskthree{}).
We then evaluate state-of-the-art foundation models (FMs) on the three benchmarks. 
Our analysis find that (i) reasoning models significantly outperform non-reasoning models on scheduling and constrained text generation tasks, their gains rising with increasing task complexity; (ii) all models struggle consistently with negation constraints; and (iii) model performance in visual causal reasoning tasks is limited by visual processing of images.

%% file: figures/tex/comparison.tex
\begin{table*}[h]
\small
\centering
\begin{tabular}{  c  c  c  c  c  c  c }
  \toprule
  \makecell{\bf Method} &
  \makecell{\bf Modality} &
  \makecell{\bf Domain} &
  \makecell{New Benchmark\\Creation} &  
  \makecell{Supports Open-Ended \\Tasks}  &
  \makecell{Controllable \\ Parameters} & 
  \makecell{Automated \\ Verification}
  
  \\ 
  \midrule
  \makecell{\citet{xia2024leaderboardrankingcoding}} & Language & Coding &  {\color{red}\faTimes} & 
  {\color{darkgreen}\faCheck} & {\color{darkgreen}\faCheck} & {\color{darkgreen}\faCheck} \\
  
  \makecell{\citet{zhang2024task}} & Multimodal & VQA &  {\color{red}\faTimes} & 
  {\color{red}\faTimes} & {\color{darkgreen}\faCheck} & N/A \\
  \makecell{\citet{wang2024benchmarkselfevolvingmultiagentframework}} & Language & General & {\color{red}\faTimes} &
  {\color{red}\faTimes} & {\color{darkgreen}\faCheck} & {\color{darkgreen}\faCheck}  \\
   \makecell{\citet{yuan2024sevalautomaticadaptivetest}} & Language & Safety & {\color{red}\faTimes} & 
   {\color{red}\faTimes} &{\color{darkgreen}\faCheck} & {\color{darkgreen}\faCheck}  \\
  
  \makecell{\citet{li2024autobenchercreatingsalientnovel}} & Language & General   & {\color{red}\faTimes}  
 & {\color{red}\faTimes} &{\color{darkgreen}\faCheck} & {\color{darkgreen}\faCheck}\\
  \makecell{\citet{zhu2024dyvaldynamicevaluationlarge}} & Language & Reasoning & {\color{darkgreen}\faCheck} 
  & {\color{red}\faTimes} & {\color{darkgreen}\faCheck} & N/A   \\

  \makecell{{\bf \sys{}}} & Multimodal & General & {\color{darkgreen}\faCheck} 
  &{\color{darkgreen}\faCheck} &{\color{darkgreen}\faCheck} & {\color{darkgreen}\faCheck} \\
  \bottomrule
\end{tabular}
\caption{\footnotesize Comparison of automated benchmark creation frameworks.}
\label{tab:table_with_icons_text_colored}
\end{table*}

%% file: sections/2_related_work.tex
\section{Related Work}
\label{sec:related}

\input{figures/tex/teaser}

\noindent {\bfseries Dynamic Benchmark Creation:}~\citet{zhang2024task} select images and scene graphs from a corpus and generate templated question-answer pairs for custom multi-modal evaluations.
~\citet{yuan2024sevalautomaticadaptivetest} propose \texttt{AutoBench} for aligning vision-language model evaluation, annotating images with question-answer pairs using LLMs for skill analysis.~\citet{zhu2024dyvaldynamicevaluationlarge} design an algorithm for evaluating reasoning tasks using graphs. These methods let users generate fine-grained evaluation data, but the generators are {\em custom-built for specific tasks, limiting broader applicability}.

\noindent {\bfseries Benchmark Extension:}~\citet{li2024autobenchercreatingsalientnovel} propose \texttt{AutoBencher}, which uses an LLM to enhance benchmark diversity by generating question-answer pairs from retrieved topic-relevant content. While effective for QA, it is {\em non-trivial to extend} to broader generative tasks. ~\citet{wang2024benchmarkselfevolvingmultiagentframework} present a multi-agent framework for dynamically augmenting benchmarks for scalability and robustness. ~\citet{xia2024leaderboardrankingcoding} look at evolving existing coding benchmarks into different coding domains using LLM-based augmentation and manual verification. Though dynamic and scalable, these approaches {\em mandatorily require a seed dataset} to bootstrap the process.  

Table~\ref{tab:table_with_icons_text_colored} summarizes our comparisons.

%% file: figures/tex/teaser.tex
\begin{figure*}
    \centering
    \includegraphics[width=1\linewidth]{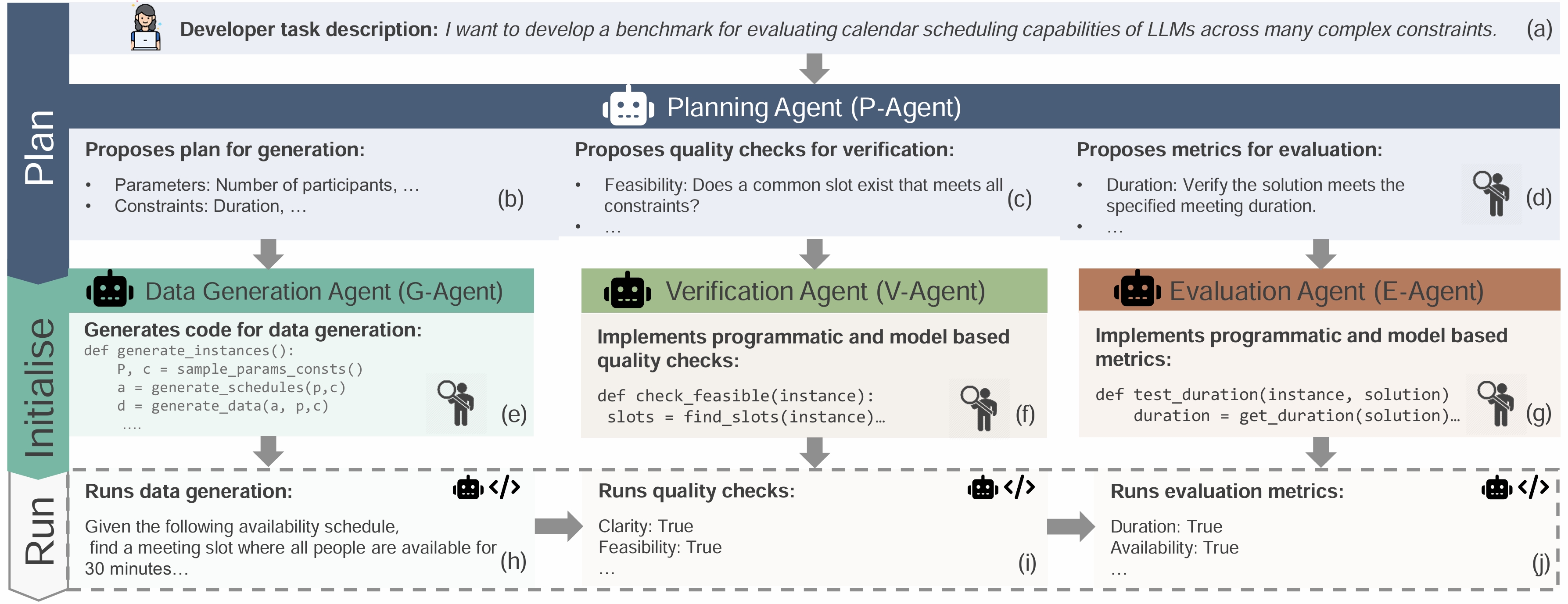}
    \caption{\footnotesize Overview of \sys{}. \pagent{} generates a plan for data generation and communicates this to \gagent{}. \gagent{} writes code for data generation and communicates to all agents. \pagent{} generates plans for evaluation and verification and communicates these to the respective agents. \eagent{} and \vagent{} write code for evaluation and verification. For each instance, generation, verification and evaluation are run.}
    \label{fig:framework}
\end{figure*}

%% file: sections/3_components.tex
\section{\sys{}}
\label{sec:details}

\input{figures/tex/dataset_overview}

\sys{} employs a multi-agent architecture to automate the benchmark creation. The framework outputs (i) diverse and verified benchmark instances and (ii) metrics and executable evaluation code to assess target model performance on these instances. Fig.~\ref{fig:framework} illustrates this workflow using a calendar scheduling task as an example.

\noindent \textbf{Planning Agent (\pagent{})} orchestrates the benchmark creation process. Its input is the developer's task specification in natural language e.g., \emph{"I want a benchmark for evaluating calendar scheduling capabilities of LLMs"} and an optional set of seed input-output pairs. It then systematically decomposes the benchmark creation process into plans for the other agents as follows:
\begin{myitemize}
    \item Plan for Data Generation (\gagent{}):
    \begin{myitemize}
        \item task-specific \emph{parameters} that can be systematically varied to create diverse evaluation instances. These parameters are used to condition instance generation and enable fine-grained control over the benchmark data distribution. Examples of parameters in calendar scheduling task include number of participants.
        \item values or ranges for parameters to ensure comprehensive and controllable coverage of the problem space (e.g., 2-10 participants). 
        \item optional constraints specifying the conditions that make a solution correct. 
        Constraints establish the success criteria for the task and determine its difficulty level. 
        In calendar scheduling, constraints might include minimum meeting duration, required buffer times between meetings, etc -- all of which must be simultaneously satisfied. 
        \item optional tools or libraries to utilize for data generation (e.g., for generating images, charts etc.)
    \end{myitemize}
    \item Plan for Verification (\vagent{}): a set of quality checks and their detailed definitions to verify the quality of each instance. These checks ensure that the generated instances are exemplar representatives of the task and that they can support reliable evaluations.
    \item Plan for Evaluation (\eagent{}): a suite of evaluation metrics to assess model performance.
\end{myitemize}

Upon plan creation, developers can further steer benchmark creation to better align with their measurement goals by refining any proposed elements.

\noindent \textbf{Generation Agent (\gagent{})} outputs executable code to generate diverse benchmark instances (Fig.~\ref{fig:framework} box e). 
For each benchmark instance, \gagent{} first samples specific parameter and constraint values to form instance metadata. Sampling structured values ensures that the distribution and coverage of the benchmark data are aligned with the plan. For cases requiring consistent format and structure, \gagent{} creates static templates that are populated with the sampled parameters. For instances needing diversity, \gagent{} creates prompt templates that instruct LLMs to generate content conditioned on the specific sampled parameter values.  To enrich the data generation \gagent{} can also incorporate tools and libraries suggested by the \pagent{}. Beyond text generation, this can enable generation of multi-modal content (images, charts, or structured metadata) when required by the benchmark specification.

\noindent \textbf{Verification Agent (\vagent{})} implements instance-level quality checks to analyze and verify generated instances. We define a general set of quality checks below: 

\begin{myitemize}
\item {\em Clarity:} The instance should be understandable/unambiguous for developers and target models. 

\item {\em Completeness:} The instance should contain all the parameters and constraints present in the sampled metadata. 
For the example in Fig.~\ref{fig:framework}, meeting duration should be present in all prompts. 

\item {\em Consistency:} When a parameter or constraint is realized in the instance, its value should be consistent with the metadata. 
For the example in Fig.~\ref{fig:framework}, the ``number of participants'' parameter should be consistent with the number of participants in the schedules.

\item {\em Feasibility:} The constraints should define a problem with at least one valid solution. 
For e.g, in Fig.~\ref{fig:framework}, a common time slot should exist that satisfies all constraints.

\item {\em Complexity:} The constraints should be associated with a measure of difficulty. To capture this, a task-specific metric is defined by \pagent{}. For the example in Fig.~\ref{fig:framework}, the metric involves the ratio of feasible slots to all slots. 
\end{myitemize}

The \vagent{} implements these checks (along with those suggested by the developer) by generating code (i.e., programmatic checks) or by generating prompts (i.e., model-based checks). 

\noindent \textbf{Evaluation Agent (\eagent{})} implements the comprehensive suite of evaluation metrics defined by \pagent{}. 
Unlike traditional benchmarks that rely on simple ground-truth comparisons, \eagent{} also addresses the challenge of evaluating open-ended generative outputs by combining programmatic evaluation and context-aware model-based evaluation. For programmatic metrics, \eagent{} generates the code for task-specific evaluation. 
For the example in Fig.~\ref{fig:framework}, \eagent{} is required to check if the solution complies with the duration constraint (box g). To achieve this, it generates code to extract and check the duration of the solution, given access to parameters and constraints (box d). 
Alternatively, for model-based evaluation, \eagent{} generates specific prompts for LLMs to judge the quality of a generated solution. 


{All generated code, verification checks, and evaluation metrics produced by the agents are verified and evaluated for correctness (details in \S~\ref{sec:quality}.)}

%% file: figures/tex/dataset_overview.tex
\begin{table*}[h!]
\scriptsize
    \centering
\renewcommand{\arraystretch}{1.5} 
    \begin{tabular}{p{2.5cm}|p{13.5cm}}
        \toprule
        \textbf{Dataset} & \textbf{Example} \\
        \midrule
        \midrule
        BA-Calendar & You are a scheduling assistant. Given the following availability schedules for participants, find an earliest common time slot for a meeting that lasts 45 minutes. Note that no meetings should be scheduled after 18:00. Availability: p1: Wednesday [09:00-14:00] - p2: Wednesday [09:00-09:45, 12:30-13:15] - p3: Wednesday [09:00-10:30, 12:30-13:15, 16:45-17:00] - p4: Wednesday [12:00-12:15, 12:30-13:15]. What is the common time slot for the meeting?  \\ 
        \hline
        BA-Text & You are an engineer tasked with writing a technical report on energy storage systems. Your report should begin with an introduction to energy storage systems and end with a conclusion. Include detailed descriptions of the types of energy storage systems, and incorporate sections on the efficiency of different energy storage technologies. If you discuss battery storage, include a section on lithium-ion batteries. 
        If you include cost analysis, compare initial investment versus long-term savings. For each type of energy storage system, provide an efficiency analysis and include a case study example. Follow the introduction with a detailed analysis of different types of energy storage technologies. \\
        \hline
        BA-Causal & 
\includegraphics[width=\linewidth]{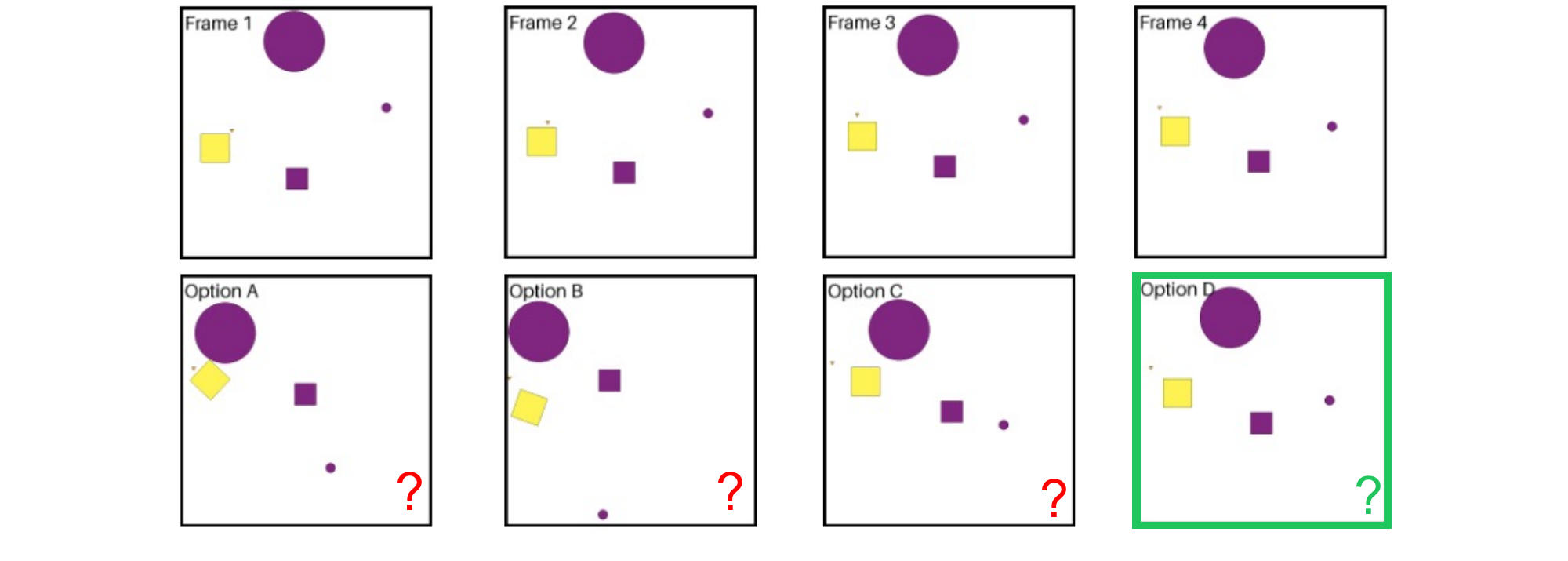}
Examine the sequence of 4 images showing the trajectory of objects in a 2D space over time steps $t_1, t_2, t_3, t_4$. Think step by step and describe the changes observed between each consecutive image. Using this information, determine the most likely next state of the system ($t_5$) from the options (a, b, c, d)\\
        \hline
    \end{tabular}
    \caption{\footnotesize Example of task instance from each Benchmark generated using \sys{}}
    \label{tab:dataset_overview}
\end{table*}

%% file: sections/4_benchmark_generation.tex
\section{Benchmark Generation}
\label{sec:generation}

We describe how we used \sys{} to generate benchmarks for calendar scheduling, constrained long-form text generation, and visual causal reasoning.

\vspace{-4pt}
\subsection{Calendar Scheduling (\taskone{})}
\label{subsec:t1}
\input{figures/tex/precision}

Calendar scheduling tests models' planning and reasoning capabilities. Previous work's~\citep{zheng2024naturalplanbenchmarkingllms} (\textsc{NaturalPlan}) limited coverage (two participants, one-day events) hinders real-world applicability. To address this, we create \taskone{}. 

As part of the plan, \pagent{} proposes (i) {\em task parameters:} number of participants, number of days with availability, days of the week, minimum/maximum length of time block, minimum/maximum number of blocks per day, earliest start time, and latest end time; and (ii) {\em constraints:} each participant’s availability, required meeting duration, buffer times, no meetings on weekends, time-of-day restrictions, priority meetings, and blackout periods.
\gagent{} generates data creation code based on these inputs.
\vagent{} performs model-based checks for clarity, completeness, and consistency, as well as programmatic feasibility checks. It also computes a task-specific complexity score based on \emph{constrainedness}: the complement of the ratio of feasible solutions to the number of time slots with at least one available participant.
Finally, \eagent{} initializes programmatic metrics for constraint satisfaction.

\subsection{Long-form Text Generation (\tasktwo{})}
\label{subsec:t2}

Constrained long-form text generation tests a model’s ability to produce responses that satisfy multiple user-specified requests—a key capability for creative and technical writing in productivity applications. Existing benchmarks focus on format-only constraints~\citep{xia2024fofobenchmarkevaluatellms}, short-form outputs~\citep{zhou2023instructionfollowingevaluationlargelanguage}, or relatively simple constraints~\citep{yao2023colliesystematicconstructionconstrained}.
In contrast, \tasktwo{} targets long-form generations with complex content-based constraints.

\pagent{} proposes (i) parameters: user, role, and task, and (ii) constraints, including:
a) \emph{Positive constraints} (inclusion of specific topics or entities),
b) \emph{Negative constraints} (exclusion of certain content),
c) \emph{Positional constraints} (placement at specific locations, e.g., paragraph),
d) \emph{Sequencing constraints} (required content order),
e) \emph{Conditional constraints} (inclusion/exclusion based on conditions), and
f) \emph{Iterative constraints} (repeated application of prior constraints).
\gagent{} generates data creation code based on these inputs.
\vagent{} performs model-based checks for clarity, completeness, consistency, and feasibility, and computes a programmatic \emph{constrainedness} score: the ratio of constraints applied to the total defined.
Finally, \eagent{} initializes model-based evaluation metrics \footnote{Since the task is open-ended and generative, we use LLM-as-judge evaluation. Metric–human alignment is discussed in Appendix~\ref{app:eval_check}.} for constraint satisfaction.

\input{figures/tex/verifier} 

\subsection{Visual Causal Reasoning (\taskthree{})}
\label{subsec:t3}

Prior visual reasoning benchmarks like CLEVR~\citep{clevr} and VQA~\citep{vqa} emphasize compositional and spatial reasoning in 2D but do not assess causality. More advanced datasets such as CLEVRER~\citep{clevrer}, CRAFT~\citep{craft}, and PHYRE~\citep{phyre} introduce causal elements via 3D or video formats, but also bring confounders like depth estimation and multi-view integration. \taskthree{} addresses this gap with a controlled 2D setup for causal reasoning. Using physics-based simulations, \sys{} generates four-step image sequences ($t_1$–$t_4$), and tasks models with predicting the correct next state ($t_5$) from one true and three distractor options, enabling clean evaluation of visual causal inference.

For this task, \pagent{} uses the \texttt{PyMunk} library~\citep{pymunk} to simulate physical environments with parameters such as gravity, number of objects, and object types/sizes. \gagent{} generates code accordingly, showcasing \sys{}’s compatibility with external tools. Each simulation runs for 20 timesteps: the first four frames ($t_1$–$t_4$) form the input, $t_5$ is selected as the ground-truth next state, and three distractors are sampled from the remaining frames.
As data is deterministically generated, clarity and consistency are guaranteed.
\emph{Task complexity is measured as one minus the normalized Euclidean distance between the answer and its closest distractor}; lower scores indicate easier instances. Instances with complexity below 0.99 are retained to ensure clear answer separability.
\eagent{} computes exact match accuracy against the ground truth.

For all tasks, we use GPT-4o (05-13) to power the various agents. 
Table~\ref{tab:dataset_overview} provides example instances; agent configurations are detailed in Appendix~\ref{app:prmpts}. Comprehensive parameters, constraints, data generation steps, and their sources (agent configuration, model, developer feedback) are in Appendices~\ref{sec:cal_bench_details},~\ref{sec:text_bench_details}, and~\ref{sec:causal_bench_details}.

%% file: figures/tex/precision.tex


\begin{table}[h]
\centering
\scriptsize
\begin{tabular}{ccc}
\toprule
{\bf Metric} & \taskone{} & \tasktwo{} \\ 
\midrule
Clarity         & 0.96 & 0.80 \\
Completeness    & 0.90 & 0.96 \\
Consistency     & 0.86 & 0.76 \\
Feasibility     & N/A  & 0.76 \\
\bottomrule
\end{tabular}
\caption{\footnotesize Accuracy for \vagent{} model-based test results with human annotated ground truths.}
\label{tab:human_verifier}
\end{table}

%% file: figures/tex/verifier.tex


\begin{table}[h]
\centering
\scriptsize
\begin{tabular}{cccc}
\toprule
{\bf Metric} & \taskone{} & \tasktwo{} & \taskthree{} \\
\midrule
Clarity         & 0.99 & 0.84 & N/A \\ 
Completeness    & 0.96 & 0.94 & N/A \\ 
Consistency     & 0.96 & 0.89 & N/A \\ 
Feasibility     & 0.93 & 0.73 & 0.99 \\ 
\bottomrule
\end{tabular}
\caption{\footnotesize Pass rate for verification quality checks.}
\label{tab:verifier}
\end{table}

%% file: sections/5_quality.tex
\section{Benchmark Quality Assessment}
\label{sec:quality}

To validate the quality and diversity of the benchmarks produced by \sys{}, we perform a quality assessment consisting of automated and human-based assessments.

\input{figures/tex/ba_calendar_aaai}

\noindent \textbf{Are Verification Checks Accurate?} For programmatic quality checks, developers can manually review the generated code to ensure correctness. Optionally, model-based quality checks may also be included, as demonstrated in \taskone{} and \tasktwo{} (but not \taskthree{}).
To evaluate the effectiveness of these checks, we conducted a human assessment study in which annotators reviewed examples and labeled quality violations. We then measured agreement between model-based checks and human annotations (see Appendix~\ref{app:human}).
Results in Table~\ref{tab:human_verifier} show that model-based checks are generally aligned with human judgments. We filter out any instance that fails \emph{any} check, making the combined verification steps a reliable quality filter. These checks can be further strengthened by providing in-context examples of low-quality or out-of-scope generations, when available.

\input{figures/tex/constrainedness_text}
\noindent \textbf{Are Generated Instances High Quality?} We evaluate the quality of instances generated by \gagent{} using two criteria: (i) compliance with verification checks described in \S~\ref{sec:details}, and (ii) coverage of key parameters relevant for comprehensive evaluation.
Table~\ref{tab:verifier} reports pass rates for each verification check. We find that \gagent{} produces high-quality instances across all tasks, supporting the robustness of our benchmarks. \taskone{} shows consistently higher quality than the more complex \tasktwo{}. Further analysis reveals that \gagent{} occasionally produces instances with conflicting constraints or inconsistent parameters, which are subsequently filtered out by \vagent{}. Since \vagent{} removes any instance that fails a verification check—programmatic or model-based—the resulting dataset is high quality.
For \taskthree{}, instances are correct by construction; only a feasibility check, as defined in \S~\ref{sec:details}, is applied—and it shows a high pass rate.
Coverage of key parameters is detailed in Appendices~\ref{app:1_coverage} and~\ref{app:2_coverage}. We observe that \sys{} generates diverse instances with broad parameter coverage, outperforming even manually curated benchmarks (see Appendix~\ref{app:1_coverage}).

\noindent \textbf{Are Generated Instances Difficult?} While we did not explicitly optimize for difficulty, a strong benchmark should present non-trivial challenges. To assess this, we follow prior work~\citep{abdin2023kitab,yuksekgonul2023attention} and evaluate whether the complexity metrics defined in \S~\ref{subsec:t1}, \S~\ref{subsec:t2}, and \S~\ref{subsec:t3} serve as reliable proxies for difficulty.
We analyze GPT-4o’s performance across difficulty levels by grouping task instances into buckets based on their complexity scores and reporting average performance and instance counts per bucket (Fig.~\ref{fig:pass_rate_vs_constraints}).
Across all tasks, we observe a monotonic drop in performance as complexity increases (for buckets with >100 instances), indicating that added constraints do indeed raise difficulty. This suggests that the generated benchmarks present a meaningful challenge even for advanced models like GPT-4o. 
To ensure that the generated benchmarks are solvable, we conduct a human baseline by asking 2 annotators to answer 20 randomly selected questions in each benchmark. Our human solvers scores 85\% for \taskone{} and 97.5\% for \taskthree{}. For \tasktwo{}, as the questions highly domain specific and intensive, human annotators took approximately 1-1.5 hours for each instance and hence producing a good baseline for this task is expensive.

\noindent \textbf{How Does Developer-in-the-Loop Contribute to Validity?}
\sys{} incorporates developer feedback at multiple stages to ensure benchmark validity. After \pagent{} generates plans for parameters, constraints, and metrics, developers review and refine these specifications to align with intended evaluation goals. Subsequently, developers verify the correctness of code and prompts generated by \gagent{} for data generation, \vagent{} for quality checks, and \eagent{} for evaluation metrics. As quantified in Appendix \ref{app:feedback}, the nature of developer interventions varies by task: \taskone{} required planning refinement for logical constraint parsing, while \taskthree{}'s well-defined physics parameters shifted developer focus toward implementation correctness and verification criteria.

%% file: figures/tex/ba_calendar_aaai.tex
\begin{figure*}[ht]
    \centering
    \begin{tabular}{ccc}
        \begin{subfigure}[b]{0.33\textwidth}
            \centering
            \includegraphics[width=\textwidth, keepaspectratio]{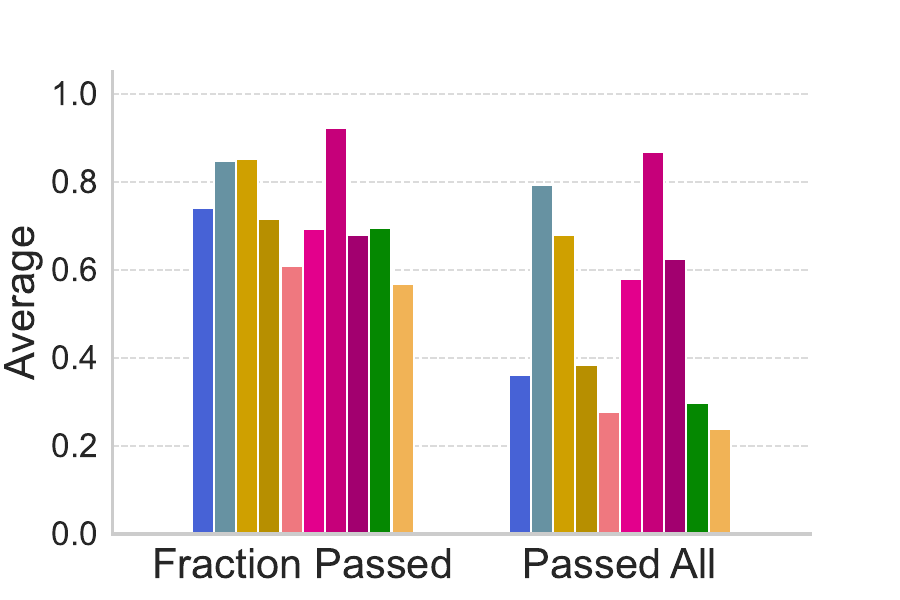}
            \caption{Performance}
            \label{fig:calendar_results}
        \end{subfigure}
        &
        \begin{subfigure}[b]{0.33\textwidth}
            \centering
            \includegraphics[width=\textwidth, keepaspectratio]{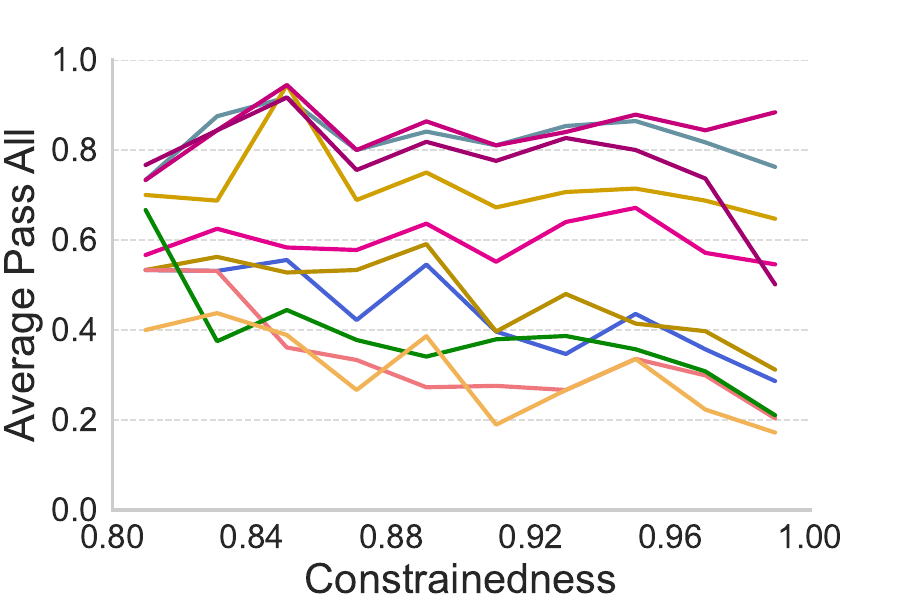}
            \caption{Accuracy v/s Constrainedness}
            \label{fig:calendar_accuracy_vs_constrainedness}
        \end{subfigure}
        &
        \begin{subfigure}[b]{0.24\textwidth}
            \centering
            \includegraphics[width=\textwidth, keepaspectratio]{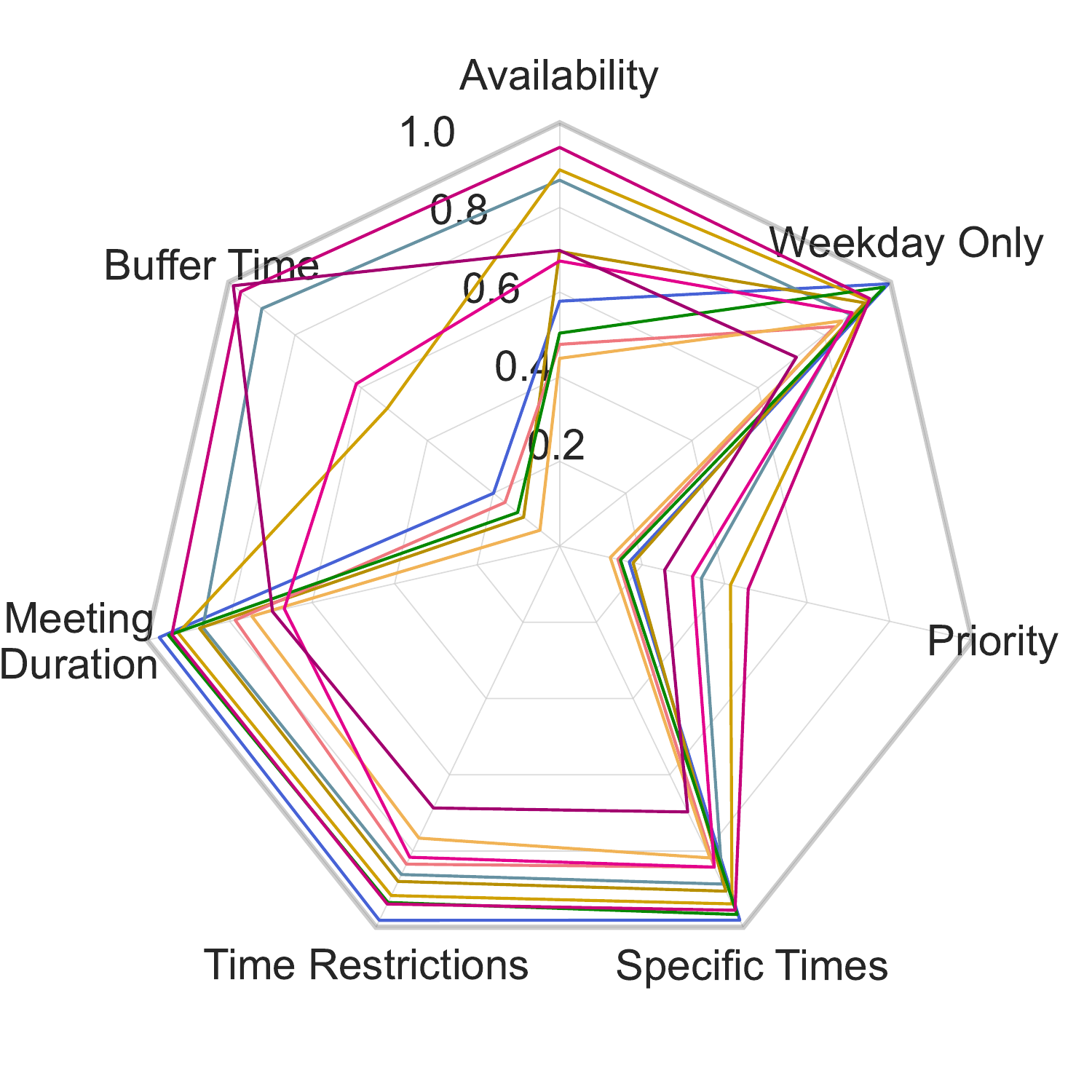}
            \caption{By Constraint}
            \label{fig:calendar_results_by_constraint}
        \end{subfigure}
    \end{tabular}
    \centering
    \begin{tabular}{c}
    \includegraphics[width=0.92\textwidth]{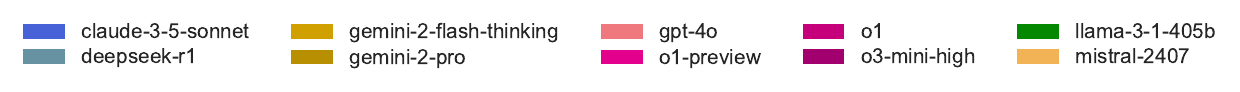} \\
    \end{tabular}
    \caption{\footnotesize \taskone{}: (a) fraction passed and pass all accuracy; (b) pass all accuracy vs. constrainedness (c) rate of `no solution' outcomes (lower is better for feasible, higher for infeasible); (d, e) pass rate for a given constraint(s) for task instances where constraints are applied}
    \label{fig:combined_evaluator_plots}
\end{figure*}

%% file: figures/tex/constrainedness_text.tex
\begin{figure*}[t] 
    \centering

    \begin{subfigure}[t]{0.33\textwidth}
        \includegraphics[width=\textwidth]{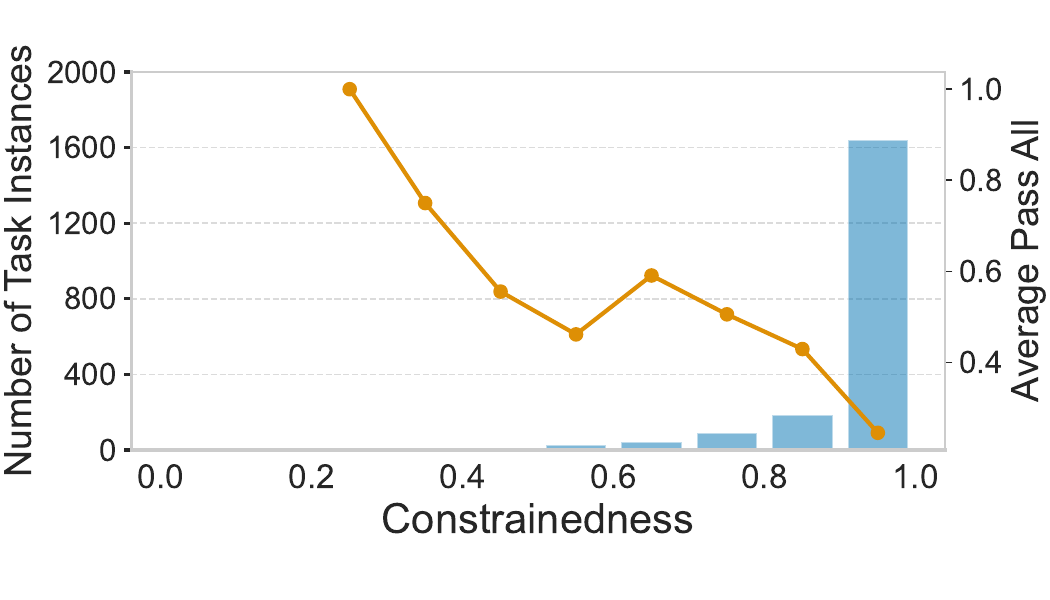}
        \caption{\taskone{}}
    \end{subfigure}\hfill
    \begin{subfigure}[t]{0.33\textwidth}
        \includegraphics[width=\textwidth]{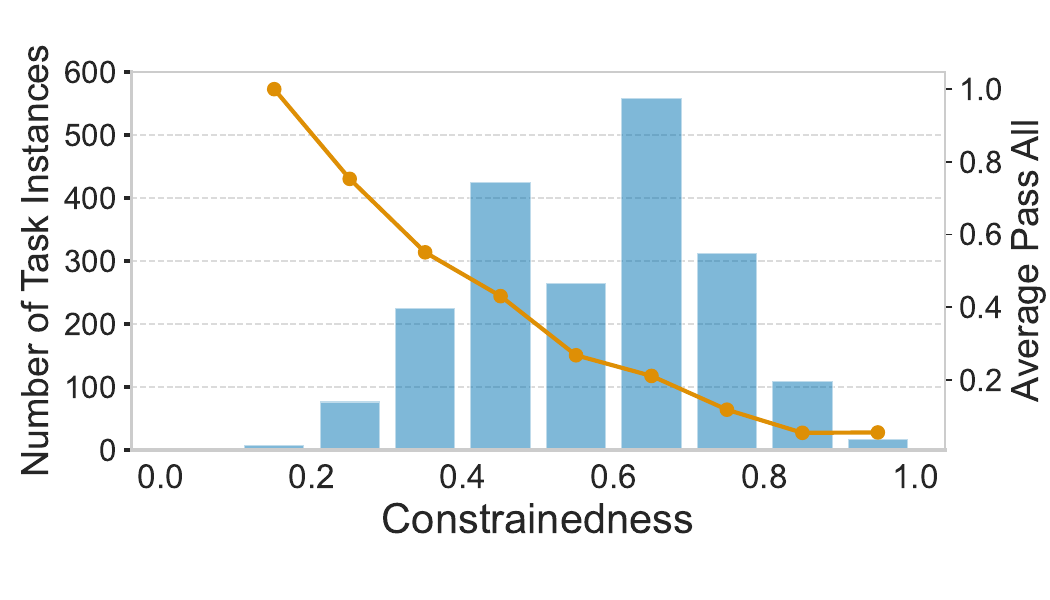}
        \caption{\tasktwo{}}
    \end{subfigure}\hfill
    \begin{subfigure}[t]{0.33\textwidth}
        \includegraphics[width=\textwidth]{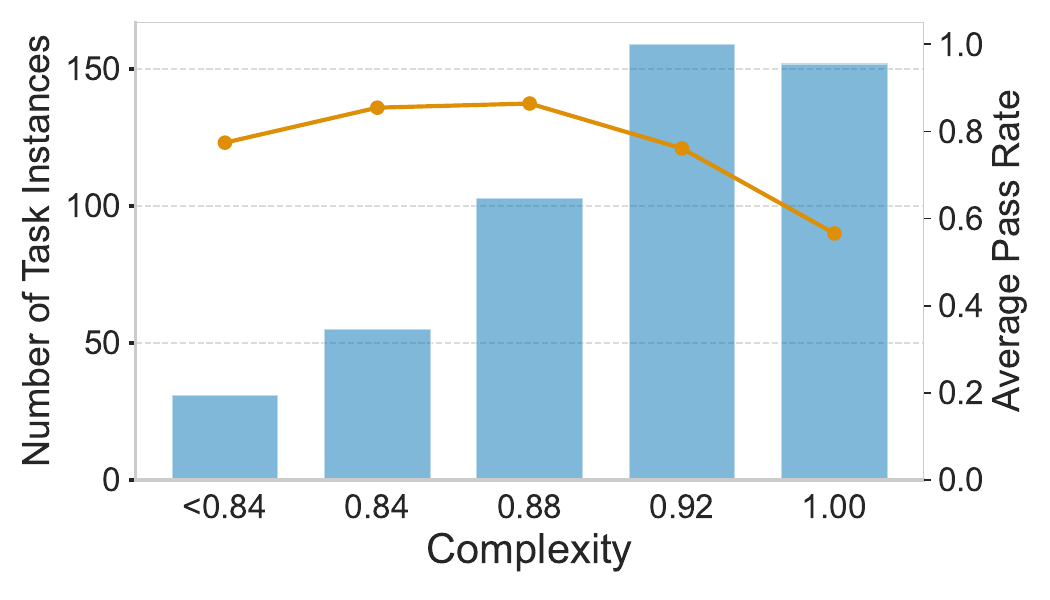}
        \caption{\taskthree{}}
    \end{subfigure}

    \caption{\footnotesize Distribution of complexity metric (bar) and GPT-4o average pass-all (line) across all tasks.}
    \label{fig:pass_rate_vs_constraints}
    \vspace{-1.0em}
\end{figure*}

%% file: sections/6_trial.tex
\section{Model Analysis}
\label{sec:final_results}

We evaluated the generated benchmarks on OpenAI o3-mini-high ~\citep{o3systemcard}, o1~\citep{o1systemcard}, GPT-4o~\citep{gpt4osystemcard}, Claude 3.5 Sonnet~\citep{ClaudeSonnet}, Gemini 1.5 Pro~\citep{reid2024gemini}, Gemini 2 Pro~\citep{Gemini2Pro}, Gemini 2 Flash Thinking~\citep{GeminiFlash}, Llama 3.1 405B~\citep{dubey2024llama}, Mistral 2407~\citep{MistralLarge2}, and DeepSeek R1~\citep{guo2025deepseek}.\footnote{Details of configurations are presented in Appendix~\ref{app:setup}.} {As evaluation overhead is substantial we chose these diverse set of models to balance representation coverage (open-source, closed, reasoning etc.) and cost and did not include recent more expensive models like Gemini 2.5 Pro or Claude 4 Sonnet.}

\input{figures/tex/ba_text_aaai}

\noindent \textbf{Task 1. Calendar Scheduling Results}

\noindent \underline{\em Performance:} We report (i) \emph{pass all accuracy}: accuracy of solutions that pass all constraints in an instance, and (ii) \emph{fraction passed}: the fraction of constraints passed per instance. Figure~\ref{fig:calendar_results} shows that the reasoning models perform well with >80\% accuracy, while non-reasoning models struggle, with less than 50\% accuracy on the task. Performance on fraction passed indicates that most models are able to satisfy >70\% of constraints, but non-reasoning models are unable to reliably satisfy all constraints, often missing a few. This showcases a strength of reasoning models: they are able to verify and re-attempt problems to reach a solutions satisfying many/all constraints. 
\\
\noindent \underline{\em Constrainedness:} Figure~\ref{fig:calendar_accuracy_vs_constrainedness} shows how different models' pass all accuracy under varying constrainedness (complexity). The results show a drop in performance for non-reasoning models and o3-mini-high as the complexity increases, showing that \emph{as the search space for solutions increases, models struggle to find the correct solution}.

\noindent \underline{\em Constraint Satisfaction:} The disaggregations in Figures~\ref{fig:calendar_results_by_constraint} (and~\ref{fig:master_calendar_pairwise_pass_rate}) show performance on specific constraints highlighting model-specific strengths and weaknesses. Reasoning models show significant improvement with respect to buffer time and priority, indicating improvements in  constraints involving more complex reasoning and arithmetic. Contrary to previous evaluations in math \cite{o3systemcard}, o3-mini-high underperforms relative to other reasoning models on the task in pass all accuracy, fraction passed, and in simpler constraints like meeting duration, often performing on-par or lower than non-reasoning models; this suggests that this \emph{distilled model even with high reasoning budget struggles to generalize to other reasoning domains}. Our analyses highlights the diversity (introduced through \sys{}) in \taskone{}, which allows analysis in various dimensions and can differentiate between specific strengths and weaknesses of the models.

\noindent \underline{\em Refusal:} In Figure~\ref{fig:master_calendar_refusal}, we report the response rate of models producing `no solution is available' for instances in \taskone{} (feasible) and task instances that were filtered out by \vagent{} for failing the feasibility test but passing all other verification tests (infeasible). We observe that o1-preview and Mistral 2407 state there is no common time slot available the most often. Furthermore, all models appear to understand the feasibility of a problem, as they respond without a solution more often on the infeasible set.\\

\input{figures/tex/ba_causal_aaai}

\noindent \textbf{Task 2. Long-form Constrained Text Generation Results} 

\noindent \underline{\em Performance:} For \tasktwo{}, we again report (i) \emph{fraction passed} and \emph{pass all}. 
In Figure~\ref{fig:text_results}, 
we observe that while reasoning models outperform their non-reasoning counterparts with respect to both \emph{pass all} and \emph{fraction passed}, their improvement is relatively small compared to \taskone{}. The gap between fraction passed and pass all accuracy is larger, indicating that all models struggle to reliably satisfy all constraints in a problem. While prior evaluations on short format-based instruction following datasets show models close to saturation~\citep{balachandran2024eureka, xia2024fofobenchmarkevaluatellms}, the best performing model on \tasktwo{} shows a pass all accuracy of 60\% \emph{showing that models still struggle to follow complex instructions when generating long text}.

\begin{figure}[t]
    \centering
    \includegraphics[width=0.33\textwidth]{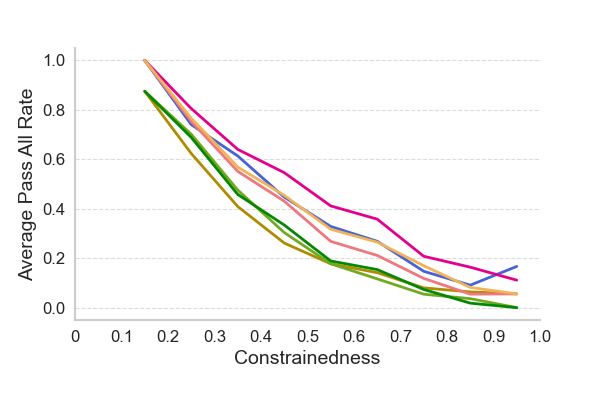}
    \begin{tabular}{c}  
    \includegraphics[width=0.4\textwidth]{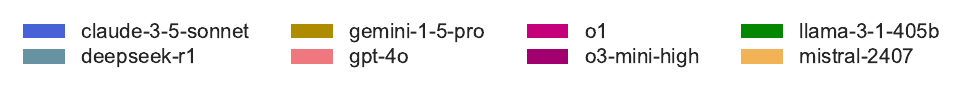} \\
\end{tabular}
    \caption{\footnotesize \tasktwo{} Accuracy v/s Constrainedness}
    \label{fig:text_passed_all_constrainedness}
\end{figure}

\noindent \underline{\em By Constraint:} Disaggregating by constraint types in Figures~\ref{fig:text_results_by_constraint} and \ref{fig:text_pairwise_pass_rate}, we see that the performance of models is stable for both positional and negative constraints. However, there is more variation with respect to iterative and conditional constraints, where DeepSeek R1, o1 and o3-mini-high perform the best. All models have similar, low performance on negative constraints, showing that they struggle with excluding specific content.
Figure~\ref{fig:text_pairwise_pass_rate} presents performance on a subset of more challenging examples that involve various constraint combinations; we observe greater variation in performance. For instance, while Llama 3.1 405B achieves a pass rate of over 60\% on tasks with only Conditional constraints, its performance drops to 20\% when both Conditional and Iterative constraints are applied. This suggests that although certain models can handle individual constraints, their reliability may diminish when additional constraints are introduced increasing task complexity.
We also observe a clustering of models in Figure~\ref{fig:text_pairwise_pass_rate} with reasoning models showing superior performance, followed by Claude 3.5 Sonnet, Mistral 2407 and GPT-4o. Finally Llama 3.1 and Gemini 1.5 Pro perform the worst on these hard examples. 
This is consistent with Figure~\ref{fig:text_passed_all_constrainedness} that shows the same order of performance, with performance gaps increase with constrainedness. 

\noindent \underline{\em Topic Consistency:} To assess topic consistency in \tasktwo{}, we conducted a focused evaluation where each model's generated response was compared to the input query to determine whether the response remained topically aligned. Specifically, we used an LLM-based evaluator to answer a binary question: \texttt{``Is the response consistent with the topic described in the query?''} For each instance, the topic constraint was isolated from other stylistic or structural constraints to avoid conflation. This binary evaluation was applied across a representative sample of instances per model. Manual spot-checking confirmed the validity of the labels. All tested models achieved average topic consistency scores above 95\%, suggesting that model responses adhered well to the intended subject matter. These results support the claim that instruction-following evaluations in \tasktwo{} are not significantly confounded by failures in factual or domain knowledge. \\

\noindent \textbf{Task 3. Visual Causal Reasoning Results}

\noindent \underline{\em Performance:} We report exact match accuracy across different dis-aggregations in our \taskthree{} dataset, observing consistent trends among the models in Figure~\ref{fig:causal_results}. GPT-4o, o1, and Gemini-2-Pro achieve the highest accuracy, while Claude-3.5-Sonnet and, in particular, Gemini-Flash-Thinking perform notably worse. 

\noindent \underline{\em Text-only Baseline:} Figure~\ref{fig:causal_text_image} compares the performance of GPT-4o and o1 when provided with inputs as images or entirely as text (details in Appendix \ref{sec:text-only}) on \taskthree{}. The results indicate that both models perform strongly when using text input, achieving near-perfect accuracy. This suggests that the reasoning required for the task is well within the capabilities of these models in a structured textual format. \emph{The drop in accuracy for image-based inputs highlights the potential challenges models face in extracting and reasoning over complex visual information, reinforcing the importance of effective alignment of vision and language}. These findings suggest that \taskthree{}, while solvable in the text domain, remains a more challenging task when approached only through visual inputs.

\noindent \underline{\em Complexity:} A breakdown by complexity in Figure~\ref{fig:causal_complexity_results} shows that all models perform best under low-complexity conditions but degrade in high complexity regime. Accuracy of GPT-4o drops from >80\% to $\sim$50\% as complexity increases. As complexity is defined by similarity between options, this highlights that \emph{models struggle when distractors closely resemble the ground-truth answer, making answer selection harder}. Interestingly, while stronger models see sharper declines in performance with increasing complexity, low performing models like Claude 3.5 Sonnet and Gemini 2 Flash Thinking show a more gradual performance decline. 

\noindent \underline{\em Varying Parameters:} In Figures~\ref{fig:causal_accuracy_vs_objects} we show performance across varying number of objects.
The top three models (GPT-4o, o1 and Gemini 2.0 Pro) demonstrate an accuracy boost as the number of objects increases, suggesting they effectively utilize the richer trajectory information. Low-performing models show a decline in performance as the number of objects increases, indicating that richer trajectory information confuses the models. This highlights that \emph{stronger models can leverage additional contextual cues from increased object interactions, whereas weaker models struggle with the added complexity, potentially due to difficulties in distinguishing relevant from irrelevant information}. This contrast suggests that the ability to integrate and reason over richer trajectory information is a key factor in model performance on this task. When evaluated on material and shape diversity (defined as the number of unique materials and shapes normalized by the total number), high-performing models show decreased accuracy (Figures~\ref{fig:causal_accuracy_vs_material}, \ref{fig:causal_accuracy_vs_shape}), indicating that increased attribute diversity introduces distractions.

%% file: figures/tex/ba_text_aaai.tex
\begin{figure*}[t]
    \centering
    \begin{tabular}{ccc}  
        \begin{subfigure}[b]{0.33\textwidth}
            \centering
            \includegraphics[width=\textwidth]{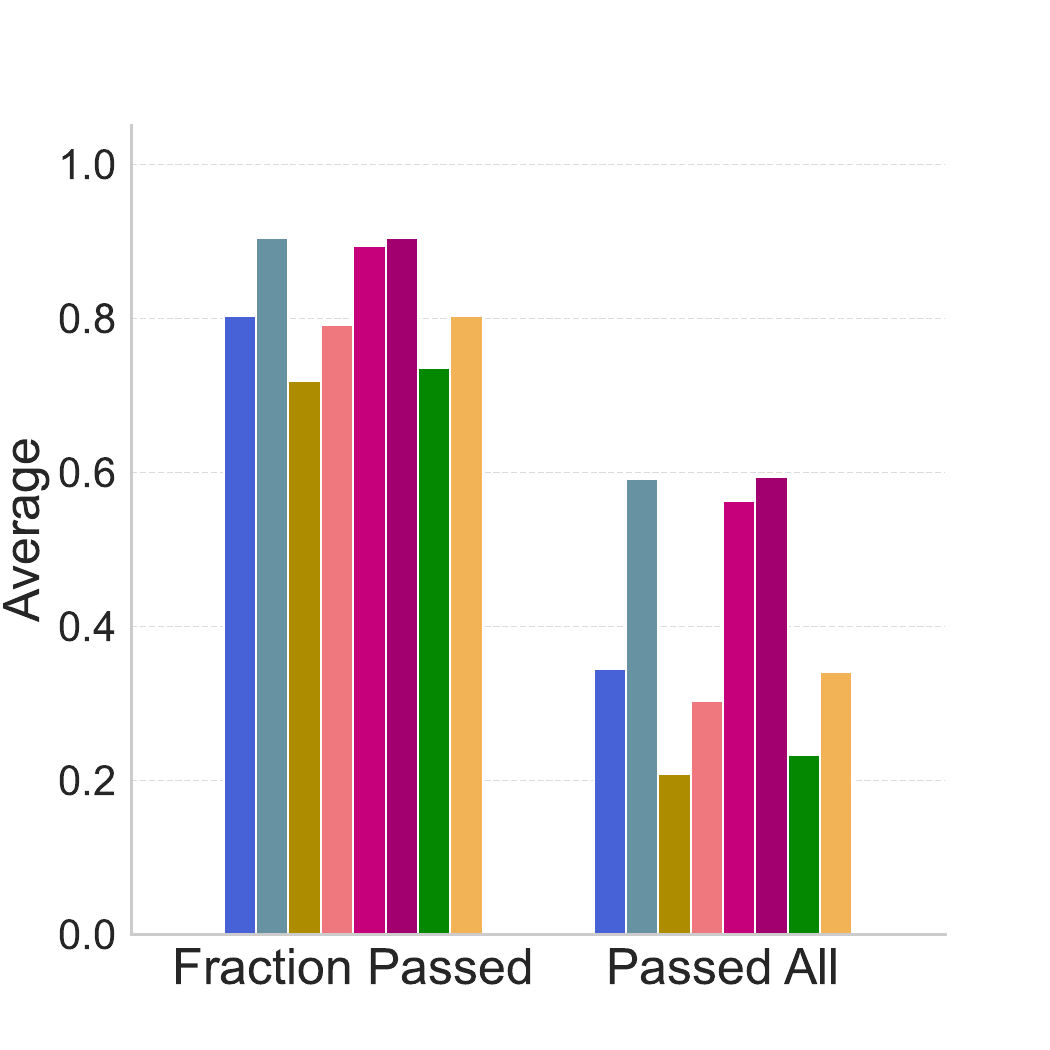}
            \captionsetup{}
            \caption{Performance}
            \label{fig:text_results}
        \end{subfigure}
        &
        \begin{subfigure}[b]{0.33\textwidth}
            \centering
            \includegraphics[width=\textwidth]{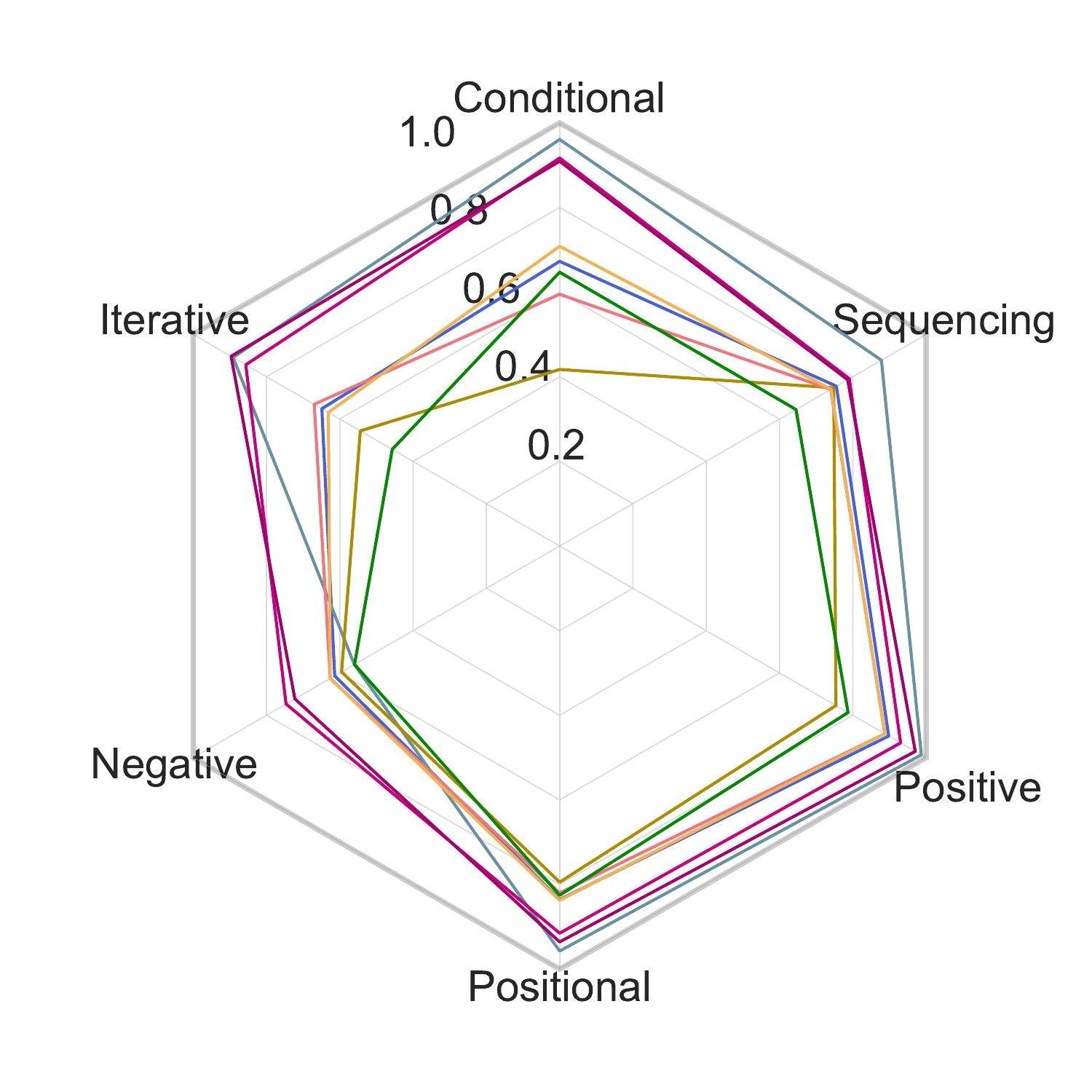}
            \captionsetup{}
            \caption{By Constraint}
            \label{fig:text_results_by_constraint}
        \end{subfigure}
        &
        \begin{subfigure}[b]{0.33\textwidth}
            \centering
            \includegraphics[width=\textwidth]{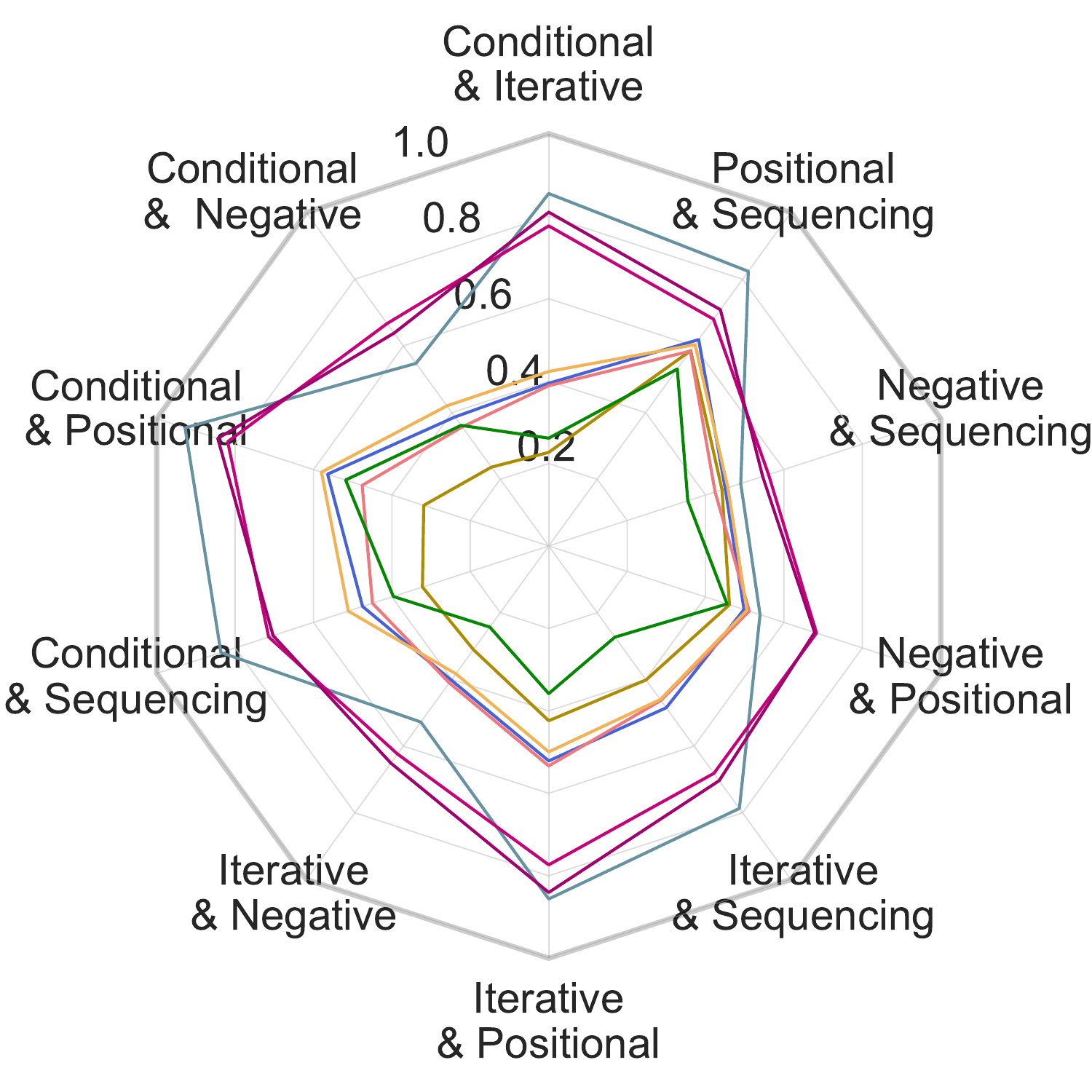}
            \captionsetup{}
            \caption{By Pairwise}
            \label{fig:text_pairwise_pass_rate}
        \end{subfigure}
    \end{tabular}
    
    \centering
    \begin{tabular}{c}  
        \includegraphics[width=0.72\textwidth]{figures/legends/legend_image_text_rows2_text_rows2.pdf} \\
    \end{tabular}
    \caption{\footnotesize \tasktwo{}: (a) shows fraction passed and pass all accuracy; (c) and (d) show pass rate for instances with specific constraint(s)}
    \label{fig:combined_evaluator_plots}
    
\end{figure*}

%% file: figures/tex/ba_causal_aaai.tex



\begin{figure*}[ht]
    \centering
    \begin{tabular}{ccc}  
        \begin{subfigure}[b]{0.33\textwidth}
            \centering
            \includegraphics[width=\linewidth]{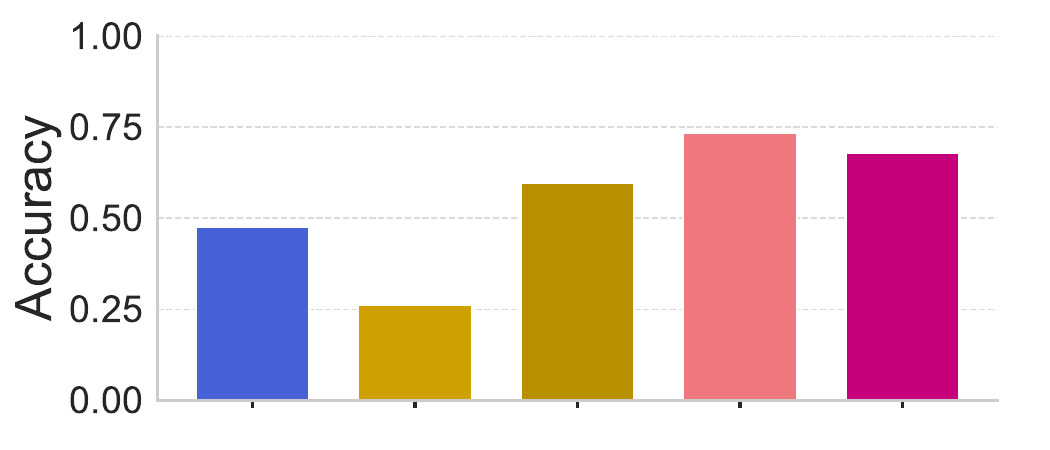}
        \captionsetup{}
        \caption{ Performance}    \label{fig:causal_results}
        \end{subfigure}
        &
        \begin{subfigure}[b]{0.33\textwidth}
            \centering
            \includegraphics[width=\textwidth]{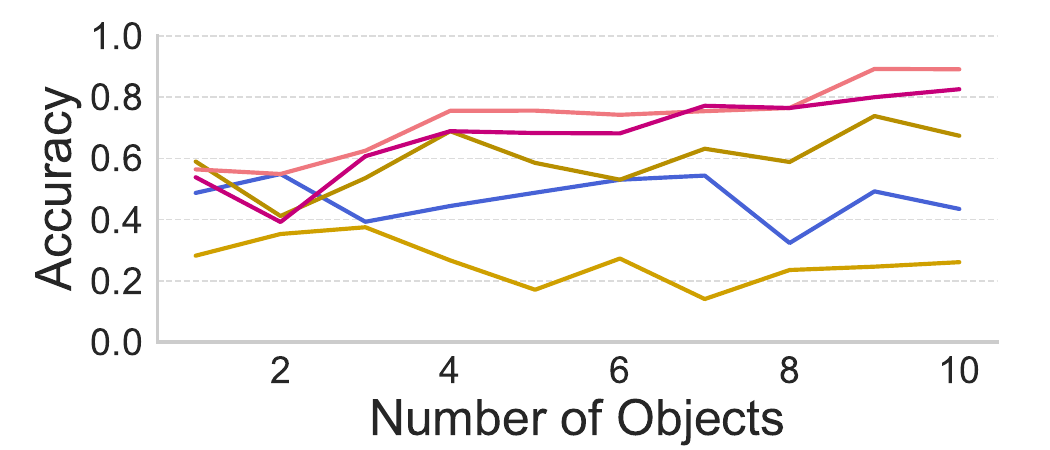}
        \captionsetup{}
        \caption{\# of Objects }
            \label{fig:causal_accuracy_vs_objects}
        \end{subfigure}
        &
    \begin{subfigure}{0.33\linewidth}
        \centering
        \includegraphics[width=\linewidth]{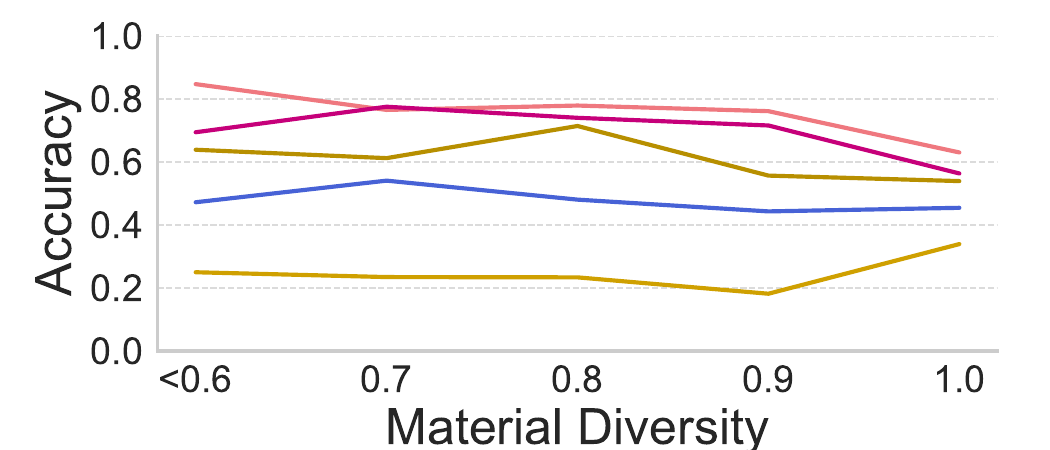}
        \caption{Material Diversity}
        \label{fig:causal_accuracy_vs_material}
    \end{subfigure}
    \\
    \begin{subfigure}[b]{0.33\textwidth}
            \centering
            \includegraphics[width=\linewidth]{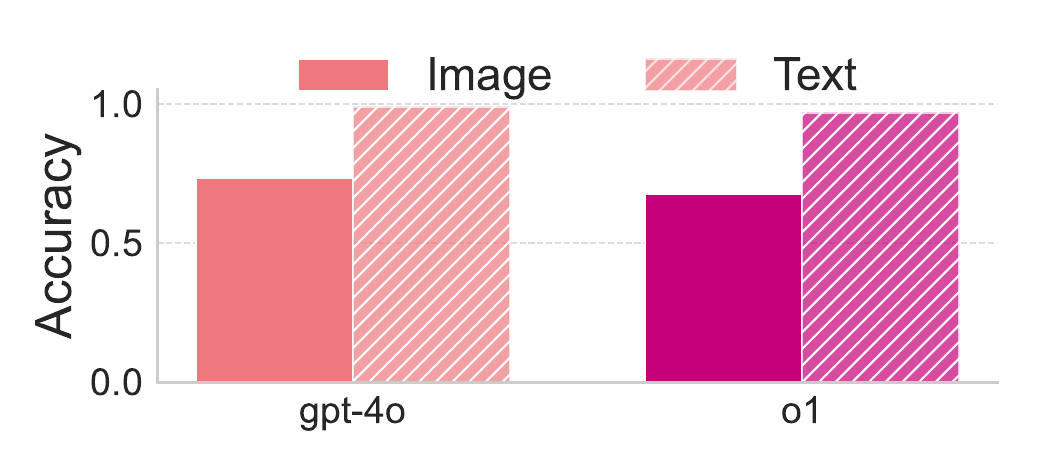}
            \captionsetup{}
            \caption{Image v/s Text Input}
            \label{fig:causal_text_image}
        \end{subfigure}
        &
        
    \begin{subfigure}{0.33\linewidth}
        \centering
        \includegraphics[width=\linewidth]{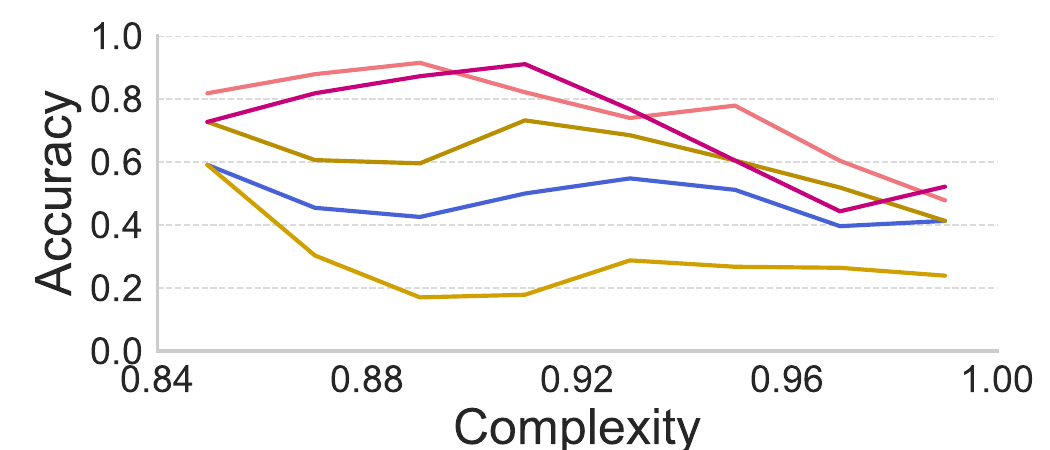}
        \caption{Accuracy v/s Complexity}
        \label{fig:causal_complexity_results}
    \end{subfigure}
    &
    \begin{subfigure}{0.33\linewidth}
        \centering
        \includegraphics[width=\linewidth]{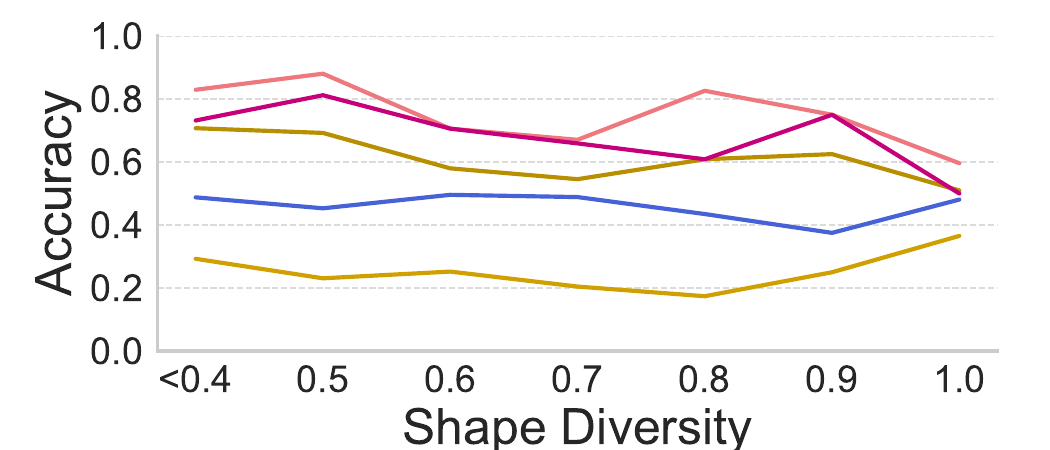}
        \caption{Shape Diversity}
        \label{fig:causal_accuracy_vs_shape}
    \end{subfigure}
    \end{tabular}
    \centering
    \begin{tabular}{c}  
        \includegraphics[width=0.85\textwidth]{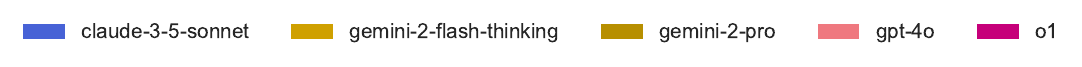} \\
    \end{tabular}
    \caption{\footnotesize \taskthree{}: (a) overall accuracy; accuracy by (b) number of objects, (c) material diversity, (e) complexity, and (f) shape diversity; (d) compares text-only vs. original image task.}
\label{fig:combined_evaluator_plots}
\end{figure*}

%% file: sections/9_conclusion.tex
\section{Conclusion}

We introduce \sys{}, a multi-agent framework for automatic benchmark creation that generates high-quality, structured, and diverse evaluations based on developer specifications, while minimizing manual effort and preserving human oversight. To showcase its utility, we use \sys{} to create three novel benchmarks targeting planning, constraint satisfaction, and visual causal reasoning, and analyze common failure modes across state-of-the-art models. \sys{}'s fine-grained evaluations reveal actionable insights: reasoning-capable models consistently outperform others on scheduling and writing tasks, making them well-suited for productivity applications—though constraints involving negation remain challenging across the board. For model developers, our findings point to visual perception, not reasoning, as the key bottleneck in causal tasks, highlighting the need for stronger multimodal grounding. More generally, \sys{} enables users to work with LLM agents to create evaluation benchmarks, reducing human effort and bias while maintaining a controllable level of human oversight.

%% file: sections/appendix/_10_appendix.tex
\input{sections/appendix/_10.1_related}
\input{sections/appendix/_10.2_generation}
\input{sections/appendix/_10.2b_generation}

\input{sections/appendix/_10.2c_generation}
\input{sections/appendix/_10.4_pass}
\input{sections/appendix/_10.5_parameter}

\input{sections/appendix/_10.6_coverage}
\input{sections/appendix/_10.3_human}

\input{sections/appendix/_10.7_settings}
\input{sections/appendix/_10.8_additional}

\input{sections/appendix/_10.9_quantifying}
\newpage
\input{sections/appendix/_10.11_full_eval}

\newpage

%% file: sections/appendix/_10.2_generation.tex
\section{More Details on Benchmark Generation 
(\S~\ref{sec:generation})}
\label{app:prmpts}

\input{figures/tex/dataset_detail_overview}

\subsection{\taskone{}}
\label{sec:cal_bench_details}

\noindent \underline{\bf Agent Configuration}
\label{sec:cal_task_desc}

Example~\ref{mybox:cal_task_desc} displays the task description for \taskone{} from the agent configuration. For the full agent configuration used for \taskone{}.

\begin{mybox}[\taskone{} Task Description from Agent Config.]
Questions will include an availability schedule for participants and additional 
constraints to increase difficulty.
The goal is to find one day and common time slot when all participants are available.
Schedules will be provided as a dictionary with participants as keys and availability 
schedules as values.
Each schedule will be a dictionary with days of the week as keys and a list of 
time blocks as values. For example:

\vspace{1em}
availability = \{
  "p1": \{
    "Monday": [9:00-12:00, 14:00-17:00],
    "Tuesday": [10:00-15:00]
  \},
  "p2": \{
    "Monday": [15:00-18:00],
    "Tuesday": [09:00-12:00, 14:00-17:00]
  \},
  "p3": \{
    "Monday": [10:00-11:00, 13:00-16:00, 18:00-19:00],
    "Tuesday": [11:00-14:00]
  \}\}
\vspace{1em}

A time block will be a string in the format "start\_time-end\_time" where a participant
is available. A time slot refers to a single time block of length meeting duration.
Granularity refers to the start and end times we consider for blocks. For example, 
if granularity is 30 minutes, we consider 9:00, 9:30, 10:00, etc.
Scheduling parameters refers to the parameters used to generate availability schedules 
such as Minium length of time block etc.
Constraints refer to additional conditions that must be met for any correct solution 
slot such as meeting duration, etc.
Constraints will always have a default value of None, False or 0.
For evaluation, solutions will be in the format:
"[day] [start\_time]-[end\_time]" or "No common time slot available".

\label{mybox:cal_task_desc}
\end{mybox}

\newpage
\noindent \underline{\bf \pagent{} Plan.}
\label{sec:cal_plan}

\begin{mybox}[\pagent{} Plan Post Developer Feedback for \taskone{}]
Parameters:\\
- Minimum length of time block: 15, 30, 45, 60 minutes\\
- Maximum length of time block: 60, 90, 120, 180, 240 minutes\\
- Number of participants: 2-10\\
- Number of days in the schedule: 1-7\\
- Maximum number of time blocks per day: 1-5\\
- Minimum number of time blocks per day: 1-5\\
- Earliest start time: 6:00, 7:00, 8:00, 9:00\\
- Latest end time: 17:00, 18:00, 19:00, 20:00\\[1em]
Constraints:\\
- Meeting duration: 15, 30, 45, 60, 90, 120 minutes\\
- Buffer time before and after meeting: None, 5, 10, 15, 30 minutes\\
- No meetings on weekends: True, False\\
- No meetings before: None, 8:00, 9:00, 10:00\\
- No meetings after: None, 17:00, 18:00, 19:00\\
- High priority meeting (must be scheduled in first available slot): True, False\\
- No meetings during specific times: None, 12:00-13:00, 16:00-17:00
\label{mybox:pagent_plan_cal}
\end{mybox}

\noindent \underline{\bf \taskone{}: Parameters \& Constraints.}

\begin{table}[h!]
\footnotesize
\centering
\begin{tabular}{ccc}
\toprule
{\bf Parameter} & Description & Provided By \\
\midrule
\midrule
Min Length Time Blocks         & Minimum length of time blocks per participant per day                 & Agent Config  \\
Number of Days         & Number of days in availability schedule                 & Model                        \\
Number of Participants         & Number of participants in availability schedule                 & Model     \\
Max Length Time Blocks         & Maximum length of time blocks per participant per day                 & Model \\
Earliest start time         & Earliest time in availability schedule                 & Model     \\
Latest end time         & Latest time in availability schedule                 & Model     \\
Min Number Time Blocks         & Minimum number of time blocks per participant per day                 & Developer\\
Max Number Time Blocks         & Maximum number of time blocks per participant per day                 & Developer\\
\bottomrule
\end{tabular}
\caption{Parameters for \taskone{}.}
\label{tab:cal_parameters}
\end{table}

\begin{table}[ht]
\footnotesize
\centering
\begin{tabular}{ccc}
\toprule
{\bf Constraint} & Description & Provided By \\
\midrule
Meeting Duration         & Meeting duration                & Agent Config \\
Availability         & Participants available                & Agent Config \\
Buffer Time         & Buffer before/after meeting                & Agent Config \\
Weekday Only & Meeting on a weekday & Model \\
Morning Time Restriction & Starts after specified time & Model \\
Evening Time Restriction & Ends before specified time & Model \\
Priority & First available slot & Developer \\
Specific Times & Exclude specific times & Developer \\
\bottomrule
\end{tabular}
\caption{Constraints for \taskone{}.}
\label{tab:cal_constraints}
\end{table}

\noindent \underline{\bf Data Generation with \gagent{}.}
As described in \S~\ref{sec:details}, \pagent{} takes the agent configuration (see Appendix~\ref{sec:cal_task_desc}) as input and generates a plan for data generation (see Appendix~\ref{sec:cal_plan}). Note, that the developer may update the plan. The plan includes a list of parameters and constraints, along with the range of values each may take. Tables~\ref{tab:cal_parameters} and~\ref{tab:cal_constraints} contains the full list of parameters and constraints from the plan along with whether each was provided by the agent configuration, LLM or developer feedback.

The \gagent{} takes the agent configuration and data generation plan as input and writes the code to initialise the data generation procedure. The data generation procedure is initialised as follows. First, the \gagent{} writes functions, \texttt{sample\_parameters()} and \texttt{sample\_constraints(parameters)}, for sampling parameters and constraints. Then, the \gagent{} writes a function \texttt{sample\_answer(constraints)} for sampling a candidate answer given the constraints. Next, the \gagent{} writes a function \texttt{sample\_availability(parameters, candidate\_answer)} for sampling availability schedules given the parameters and candidate answer. 

During data generation, \gagent{} first runs the code for sampling parameters and constraints. Then, \gagent{} sequentially runs the code for sampling a candidate answer and availability. Finally, the prompt is generated  by \gagent{} with constraints and availability schedules in-context. The instance is a triple (\texttt{prompt, parameters, constraints}). This procedure is repeated for every instance in the dataset. After filtering with \vagent{}, we have 2,000 verified instances in \taskone{}.

\begin{mybox}[\taskone{} Example Prompt from Generated Instance]
Given the following availability schedules for participants, find a common time slot for a meeting that lasts 60 minutes. Additionally, ensure there is a buffer time of 5 minutes before and after the meeting.\\[1em]
Availability:\\
p1:\\
Monday: 07:00-08:30, 09:30-12:30, 13:15-14:00, 15:00-17:15, 17:45-18:00\\
Tuesday: 07:00-11:45, 12:15-16:00, 16:45-18:00\\
Wednesday: 07:00-09:15, 09:45-11:45, 12:30-18:00\\[1em]
p2:\\
Monday: 07:00-07:15, 08:15-12:30, 13:15-15:00, 16:00-18:00\\
Tuesday: 07:00-10:15, 10:45-16:00, 17:00-18:00\\
Wednesday: 07:00-08:00, 09:00-12:30, 13:30-18:00\\[1em]
What is the common time slot for the meeting?
\label{mybox:cal_example_prompt}
\end{mybox}

\noindent \underline{\bf \taskone{}: Quality Checks \& Evaluation Metrics}
Tables~\ref{tab:cal_quality_checks} and~\ref{tab:cal_eval_metrics} contains the full list of quality checks and evaluation metrics from the \pagent{} along with whether each was provided by the agent configuration, LLM or developer feedback.
\begin{table}[ht]
\footnotesize
\centering
\begin{tabular}{ccc}
\toprule
{\bf Quality Check} & Type & Provided By \\
\midrule
\midrule
Clarity         & Model-based                 & Developer                        \\
Consistency         & Model-based                 & Model                        \\
Constrainedness         & Programmatic                 & Agent Config                        \\
Completeness         & Model-based                 & Agent Config                        \\
Feasibility         & Programmatic                 & Agent Config                        \\
\bottomrule
\end{tabular}
\caption{Quality Checks for \taskone{}.}
\label{tab:cal_quality_checks}
\end{table}

\begin{table}[ht]
\footnotesize
\centering
\begin{tabular}{ccc}
\toprule
{\bf Evaluation Metric} & Type & Provided By \\
\midrule
\midrule
Availability        & Programmatic                 & Agent Config                        \\
Meeting Duration        & Programmatic                 & Model                        \\
Buffer Time        & Programmatic                 & Model                        \\
Weekdays Only       & Programmatic                 & Model                        \\
Time Restrictions        & Programmatic                 & Model                        \\
Priority        & Programmatic                 & Model                        \\
Specific Times        & Programmatic                 & Model                        \\

\bottomrule
\end{tabular}
\caption{Quality Checks for \taskone{}.}
\label{tab:cal_eval_metrics}
\end{table}

%% file: figures/tex/dataset_detail_overview.tex
\begin{table*}[ht!]
    \centering
    \footnotesize
    \begin{tabular}{p{2cm} p{3.2cm} p{3.2cm} p{2.8cm} p{2.8cm}}
        \toprule
        \textbf{Benchmark} & \textbf{Task Description} & \textbf{Key Parameters} & \textbf{Complexity Metric} & \textbf{Evaluation Metric} \\
        \midrule
        Calendar Scheduling (\taskone{}) & Tests planning and reasoning in calendar scheduling, expanding coverage beyond prior work. & Participants, availability, meeting duration, buffer times, priority meetings. & 1 - ratio of feasible solutions to available time slots. & Constraint satisfaction metrics. \\
        \midrule
        Constrained Long-form Text Generation (\tasktwo{}) & Generates long-form text with complex constraints, surpassing prior benchmarks in depth. & User role, positive/ negative/ positional/ sequencing/ conditional constraints. & Ratio of constraints applied to total constraints. & LLM-as-judge based constraint satisfaction evaluation. \\
        \midrule
        Visual Causal Reasoning (\taskthree{}) & Evaluates causal reasoning using 2D physical simulations with counterfactual answer choices. & Gravity, number of objects, object types/sizes, elasticity, friction, mass, velocity, color & 1 - min normalized avg. Euclidean distance between answer and distractors & Exact match accuracy with ground truth. \\
        \bottomrule
    \end{tabular}
    \caption{Overview of Benchmarks}
    \label{tab:benchmark_overview}
\end{table*}

%% file: sections/appendix/_10.2b_generation.tex
\newpage
\subsection{\tasktwo{}}
\label{sec:text_bench_details}

\noindent \underline{\bf Agent Configuration.}
\label{sec:text_task_desc}
Example \ref{mybox:text_task_desc} shows the task description for \tasktwo{} from the agent configuration. For the full agent configuration used for \tasktwo{}.

\begin{mybox}[\tasktwo{} Task Description from Agent Config.]
\small
The task is open-ended text generation under constraints. The goal is to generate text that meets the constraints. Here are some settings in which we will base our task instances:
\\[1em]
\textbf{Chef}\\
Roles: amateur, professional\\
Task: recipe

\textbf{Teacher}\\
Roles: primary school, high school\\
Task: lesson plan

\textbf{Student}\\
Roles: high school, university\\
Task: essay

\textbf{Researcher}\\
Roles: machine learning, biology\\
Task: scientific article

\textbf{Engineer}\\
Roles: mechanical, structural\\
Task: technical report
\\[1em]
We will add constraints to each setting, grouped into categories. Here are the definitions of each category:\\[1em]
\textbf{Positive}: include something in the text generation\\
\textbf{Negative}: exclude something in the text generation\\
\textbf{Positional}: place a constraint at an absolute or relative position in the text\\
\textbf{Sequencing}: apply multiple constraints in a specific order\\
\textbf{Iterative}: apply a constraint repeatedly for list items\\
\textbf{Conditional}: if a condition is met, apply a constraint; otherwise, apply another or do nothing. (Avoid ambiguous phrasing such as “if condition apply constraint or another constraint”.)
\\[1em]
All constraints should be conditioned on the model context only and not on outside knowledge. No constraints should invoke outside knowledge, ask for figures, or other non-text outputs. They must be based solely on the model's context and the given setting.
\label{mybox:text_task_desc}
\end{mybox}
\newpage
\noindent \underline{\bf \pagent{} Plan.}
\label{sec:text_plan}

\begin{mybox}[\pagent{} Plan Post Developer Feedback for \tasktwo{}]
\footnotesize
\textbf{Chef}\\
**Roles: amateur, professional**\\
**Task: recipe**\\
- **Positive:** Include at least one type of fresh herb.\\
- **Negative:** Do not include any nuts.\\
- **Positional:** Add salt and pepper after all other ingredients have been mixed.\\
- **Conditional:** If the recipe includes meat, ensure it is marinated for at least 30 minutes; else, if the recipe is vegetarian, include a protein source like beans or tofu.\\
- **Sequencing:** Start with washing and chopping vegetables, then proceed to cooking.\\
- **Iterative:** For each step, specify the cooking time and temperature.\\[1em]
\textbf{Teacher}\\
**Roles: primary school, high school**\\
**Task: lesson plan**\\
- **Positive:** Include a hands-on activity.\\
- **Negative:** Do not use any digital tools.\\
- **Positional:** Introduce the main topic within the first 10 minutes of the lesson.\\
- **Conditional:** If the lesson is for primary school, include a story; else, if the lesson is for high school, include a case study.\\
- **Sequencing:** Begin with a warm-up exercise, followed by the main lesson, and end with a review.\\
- **Iterative:** For each section of the lesson, provide an estimated duration.\\[1em]
\textbf{Student}\\
**Roles: high school, university**\\
**Task: essay**\\
- **Positive:** Include at least three references to academic sources.\\
- **Negative:** Do not use first-person pronouns.\\
- **Positional:** State the thesis statement in the first paragraph.\\
- **Conditional:** If the essay is argumentative, include a counterargument; else, if the essay is descriptive, include detailed descriptions.\\
- **Sequencing:** Start with an introduction, followed by body paragraphs, and end with a conclusion.\\
- **Iterative:** For each body paragraph, include a topic sentence and supporting evidence.\\[1em]
\textbf{Researcher}\\
**Roles: machine learning, biology**\\
**Task: scientific article**\\
- **Positive:** Include a section on related work.\\
- **Negative:** Do not use informal language.\\
- **Positional:** Place the abstract at the beginning of the article.\\
- **Conditional:** If the research includes experiments, provide a detailed methodology; else, if it is a review, include a comprehensive literature survey.\\
- **Sequencing:** Start with an introduction, followed by methods, results, and discussion.\\
- **Iterative:** For each figure or table mentioned, provide a brief description in the text.\\[1em]
\textbf{Engineer}\\
**Roles: mechanical, structural**\\
**Task: technical report**\\
- **Positive:** Include a section on safety considerations.\\
- **Negative:** Do not include any speculative statements.\\
- **Positional:** Place the executive summary at the beginning of the report.\\
- **Conditional:** If the report includes calculations, provide detailed steps.\\
- **Sequencing:** Start with an introduction, followed by methodology, results, and conclusions.\\
- **Iterative:** For each section, include a summary at the end.
\label{mybox:pagent_plan_text}
\end{mybox}

\noindent \underline{\bf Data Generation with \gagent{}.}
\pagent{} takes the agent configuration (see Appendix \ref{sec:text_task_desc}) as input and generates a plan for data generation (see Appendix \ref{sec:text_plan}). For each \emph{user} parameter and constraint group defined in the task description from the agent configuration, the plan includes an example constraint. The developer has the option to update these to align the constraint generation with developer preferences, however, we can see from Table \ref{tab: edit_distance}, the normalized edit distance for \tasktwo{} data generation plan is 0.02 and so minimal changes were made. Definitions of parameters (all provided by the agent configuration) may be found in Appendix Table \ref{tab:text_parameters}.

Similarly to \taskone{}, the \gagent{} takes the agent configuration and data generation plan as input and writes the code to initialise the data generation procedure. The data generation procedure is initialised as follows. First, the \gagent{} writes a function \texttt{sample\_parameters} for parameter sampling. Note that the parameters here include a number of constraints for each constraint group. Next, the \gagent{} writes a function \texttt{generate\_topic(parameters)} which prompts the LLM to generate a topic grounded on the instance parameters. Then, the \gagent{} writes a function \texttt{generate\_constraints(parameters, topic)}, which prompts a LLM to sequentially generate constraints with all previous constraints in-context. 

During data generation, the \gagent{} first samples parameters and then generates a topic. Next, the \gagent{} sequentially generates constraints given the parameters and topic, keeping all previously generated constraints for the instance in-context. Finally, finally, the prompt is generated  by \gagent{} with constraints and topic in-context. An example of a constraint for each constraint group may be found in Table \ref{tab:text_constraints}. Note that often the number of constraints for each constraint group is more than one i.e. \gagent{} will generate multiple constraints in the same constraint group. The instance is a triple \texttt{(prompt, parameters, constraints)}. This procedure is repeated for every instance in the dataset. After filtering with \vagent{}, we have 2,000 verified instances in \tasktwo{}.

\begin{mybox}[\tasktwo{} Example Prompt from Generated Instance]
\footnotesize
You are tasked with writing a scientific article on the topic of underfitting in machine learning. \\
The article should include a detailed explanation of underfitting, provide at least one example of a model that commonly experiences underfitting, and discuss methods to mitigate underfitting in machine learning models. Ensure that you do not include any references to specific datasets, avoid mentioning any proprietary machine learning frameworks, and exclude any discussion of overfitting. \\[1em]
If the article discusses linear regression, include a section on the limitations of linear models in complex datasets. If neural networks are mentioned, provide an example of underfitting in a deep learning context. If the article includes a section on data preprocessing, discuss how insufficient data preprocessing can lead to underfitting. \\[1em]
Additionally, discuss at least three different techniques to address underfitting and provide multiple examples of underfitting in various machine learning algorithms.”
\label{mybox:text_example_prompt}
\end{mybox}

\noindent \underline{\bf \tasktwo{}: Parameters, Constraints, Quality Checks \& Evaluation Metrics.}

\begin{table}[ht]
\footnotesize
\centering
\begin{tabular}{ccc}
\toprule
{\bf Parameter} & Description & Provided By \\
\midrule
\midrule
User & User performing task & Agent Config \\
Role & Role of user & Agent Config \\
Task & Text generation task & Agent Config \\
Number of Constraints & The number of constraints applied in each constraint group & Agent Config \\
\bottomrule
\end{tabular}
\caption{Parameters for \tasktwo{}.}
\label{tab:text_parameters}
\end{table}

\begin{table}[ht]
\footnotesize
\centering
\begin{tabular}{ccc}
\toprule
{\bf Constraint Group} & Example & Provided By \\
\midrule
\midrule
Positive & Include at least one type of fresh herb & Model \\
Negative & Do not include any nuts & Model \\
Positional & Add salt and pepper after all other ingredients have been mixed & Model \\
Sequencing & Start with washing and chopping vegetables, then proceed to cooking & Model \\
Conditional & If the recipe includes meat, ensure it is marinated for at least 30 minutes & Model \\
Iterative & For each step, specify the cooking time and temperature & Model \\
\bottomrule
\end{tabular}
\caption{Example Constraints for \tasktwo{}.}
\label{tab:text_constraints}
\end{table}

Tables~\ref{tab:text_quality_checks} and~\ref{tab:text_eval_metrics} contain the full list of quality checks and evaluation metrics from the \pagent{} along with whether each was provided by the agent configuration, LLM, or developer feedback.

\begin{table}[ht]
\footnotesize
\centering
\begin{minipage}{0.40\linewidth}
\centering
\begin{tabular}{ccc}
\toprule
{\bf Quality Check} & Type & Provided By \\
\midrule
\midrule
Clarity         & Model-based    & Model        \\
Consistency     & Model-based    & Agent Config \\
Constrainedness & Programmatic   & Agent Config \\
Completeness    & Model-based    & Agent Config \\
Feasibility     & Model-based    & Model        \\
\bottomrule
\end{tabular}
\caption{Quality Checks for \tasktwo{}.}
\label{tab:text_quality_checks}
\end{minipage}
\hfill
\begin{minipage}{0.5\linewidth}
\centering
\begin{tabular}{ccc}
\toprule
{\bf Evaluation Metric} & Type & Provided By \\
\midrule
\midrule
Topic Consistency        & Model-based & Agent Config \\
Conditional Constraints  & Model-based & Agent Config \\
Positive Constraints     & Model-based & Model        \\
Negative Constraints     & Model-based & Model        \\
Positional Constraints   & Model-based & Model        \\
Sequencing Constraints   & Model-based & Model        \\
Iterative Constraints    & Model-based & Model        \\
\bottomrule
\end{tabular}
\caption{Evaluation Metrics for \tasktwo{}.}
\label{tab:text_eval_metrics}
\end{minipage}
\end{table}

%% file: sections/appendix/_10.2c_generation.tex
\newpage
\subsection{\taskthree{}}
\label{sec:causal_bench_details}

\noindent \underline{\bf Agent Configuration.}
\label{sec:causal_task_desc}
Example ~\ref{mybox:causal_task_desc} shows the task description for \taskthree{} from the agent configuration. For the full agent configuration used for \taskthree{}.

\begin{mybox}[\taskthree{} Task Description from Agent Config.]
\footnotesize
This dataset evaluates vision models on causal reasoning in dynamic 2D environments. Each example consists of: \\[1em]
- **Input:** A sequence of 4 images generated using Pymunk-based physical simulations, showing the trajectory of objects in a 2D space over consecutive time steps t1, t2, t3, t4. \\[1em]
- **Goal:** Predict the correct next state of the system (t5) from four choices (a, b, c, d). One choice is the true next state, while the others are counterfactuals sampled from earlier trajectory states or hypothetical system variations. \\[1em]
Simulations vary in gravity, object properties (mass, size, shape), initial velocities, and interaction dynamics. The dataset challenges models to identify causal consistency in visually complex, physics-based scenarios.
\label{mybox:causal_task_desc}
\end{mybox}

\newpage

\noindent \underline{\bf \pagent{} Plan.}
\label{sec:causal_plan}

\begin{mybox}[\pagent{} Plan Post Developer Feedback for \taskthree{}]
\footnotesize
\textbf{1. Space-Level Parameters}\\[1em]
\textbf{Gravity}\\
- \textbf{Magnitude Range:} 0 to 10 m/s². This range allows for simulations with no gravity (0) to Earth-like gravity (~9.81 m/s²) and slightly beyond for added variability.\\
- \textbf{Direction:} Gravity can be applied in any direction. For simplicity, consider using two components: \\
\quad - \textbf{X Component Range:} -5 to 5 m/s². Negative values simulate gravity pulling to the left, positive values to the right.\\
\quad - \textbf{Y Component Range:} -10 to 0 m/s². Negative values simulate gravity pulling downwards which is typical; positive values are not considered to keep the simulations realistic.\\[1em]
\textbf{Space Dimensions}\\
- \textbf{Fixed Width and Height:} For consistency, fix the width and height of the simulation area. A common choice might be 800 pixels wide by 800 pixels high, which provides ample space for object interactions without being too large for efficient computation and visualization.\\[1em]
\textbf{Boundary Conditions}\\
- \textbf{Friction Values:} 0 to 1. This range goes from perfectly smooth (0) to very rough surfaces (1), affecting how objects slide against boundaries.\\
- \textbf{Elasticity Values:} 0 to 1. This range allows for perfectly inelastic collisions (0, objects do not bounce) to perfectly elastic collisions (1, objects bounce without losing energy).\\[1em]
\textbf{2. Object-Level Parameters}\\[1em]
\textbf{Object Types}\\
- Circles, squares, rectangles, and equilateral triangles.\\[1em]
\textbf{Object Dimensions}\\
- \textbf{Circles:} Radius range from 5 to 50 pixels.\\
- \textbf{Squares and Rectangles:} Side lengths range from 10 to 100 pixels. For rectangles, allow for varying width and height within this range.\\
- \textbf{Equilateral Triangles:} Side length range from 10 to 100 pixels.\\[1em]
\textbf{Physical Properties}\\
- \textbf{Mass:} 1 to 10 kg. This range allows for light to moderately heavy objects, affecting how they interact.\\
- \textbf{Friction:} 0 to 1. Similar to boundary conditions, this affects how objects slide against each other.\\
- \textbf{Elasticity:} 0 to 1. Determines how bouncy the objects are upon collision.\\[1em]
\textbf{Initial Positions and Velocities}\\
- \textbf{Positions:} Randomly assign within the space dimensions, ensuring objects do not overlap at the start.\\
- \textbf{Velocities:} Range from -100 to 100 pixels per second for both X and Y components. Negative values indicate movement to the left/up, and positive values to the right/down.\\[1em]
\textbf{Colours}\\
Specify 7 distinct colours for visual differentiation. Example RGB values could be:\\
- Red: (255, 0, 0)\\
- Green: (0, 255, 0)\\
- Blue: (0, 0, 255)\\
- Yellow: (255, 255, 0)\\
- Purple: (128, 0, 128)\\
- Orange: (255, 165, 0)\\
- Cyan: (0, 255, 255)
\label{mybox:pagent_plan_causal}
\end{mybox}

\noindent \underline{\bf Data Generation with \gagent{}.}
\pagent{} takes the agent configuration (see Appendix \ref{sec:causal_task_desc}) as input and generates a plan for data generation (see Appendix \ref{sec:causal_plan}) using the PyMunk library for simulating trajectories of image frames containing multiple objects. This plan includes parameters that are to vary during the simulation and their ranges. Definitions of parameters may be found in Appendix Table \ref{tab:causal_parameters}. Note that although most of the parameters are given in the agent configuration, the parameter ranges are generated by the LLM with little developer feedback as shown in Appendix \ref{app:feedback} Table \ref{tab: edit_distance}.

The \texttt{\gagent{}} takes the agent configuration and data generation plan as input and writes the code to initialize the data generation procedure. The data generation procedure is initialized as follows. First, the \texttt{\gagent{}} writes a function \texttt{sample\_parameters} for parameter sampling. Next, the \texttt{\gagent{}} writes a function \texttt{sample\_trajectory(parameters)} which involves running a PyMunk simulation for 20 steps with initial conditions defined by the parameters from the parameter sampling. Then, the \texttt{\gagent{}} writes a function \texttt{select\_steps(trajectory)} which samples a collision step from the trajectory and uses this to define a starting frame for the input frame sequence \((t_1, t_2, t_3, t_4)\). The answer frame is \(t_5\). Three distractor frames are randomly sampled from the remaining frames in the trajectory. For the multiple-choice option frames, the answer frame and distractor frames are combined in a random order.

During data generation, the \texttt{\gagent{}} first samples parameters and then samples the trajectory. Next, the \texttt{\gagent{}} selects the input frames and option frames, recording the answer id. The instance is a triple \texttt{(inputs, options, answer\_id)}. This procedure is repeated for every instance in the dataset. After filtering with \vagent{}, we have 500 verified instances in \taskthree{}.

\noindent \underline{\bf \taskthree{}: Parameters.}

\begin{table}[ht]
\footnotesize
\centering
\begin{tabular}{lll}
\toprule
{\bf Parameter} & Description & Provided By \\
\midrule
\midrule
gravity & Gravity vector (x, y) for the simulation environment. & Agent Config \\
wall\_elasticity & Elasticity factor for the boundary walls. & Agent Config \\
wall\_friction & Friction factor for the boundary walls. & Agent Config \\
objects & List of objects. Each object includes the following properties: & Agent Config \\
\quad type & Shape type (circle, square, rectangle, triangle). & Agent Config \\
\quad size & Object size (single value, or tuple for rectangles). & Agent Config \\
\quad mass & Mass of the object. & Agent Config \\
\quad friction & Friction coefficient of the object. & Agent Config \\
\quad elasticity & Elasticity coefficient of the object. & LLM \\
\quad initial\_velocity & Initial velocity vector (x, y) of the object. & Agent Config \\
\quad color & RGB color tuple representing the object's color. & Agent Config \\
\bottomrule
\end{tabular}
\caption{Parameters for \taskthree{}.}
\label{tab:causal_parameters}
\end{table}

%% file: sections/appendix/_10.4_pass.tex
\newpage
\section{\vagent{}'s Pass Rates}
\label{app:pass}

\begin{figure}[ht]
    \centering
    \begin{subfigure}[b]{0.45\textwidth}
        \centering
        \includegraphics[width=\textwidth]{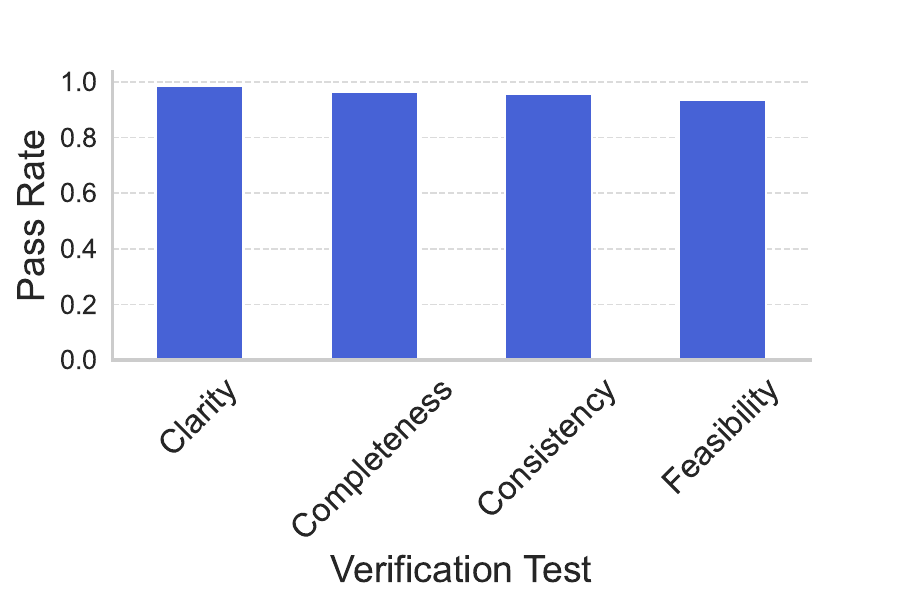}
        \caption{\taskone{}}
        \label{fig:verifier_calendar}
    \end{subfigure}
    \hspace{0.05\textwidth} 
    \begin{subfigure}[b]{0.45\textwidth}
        \centering
        \includegraphics[width=\textwidth]{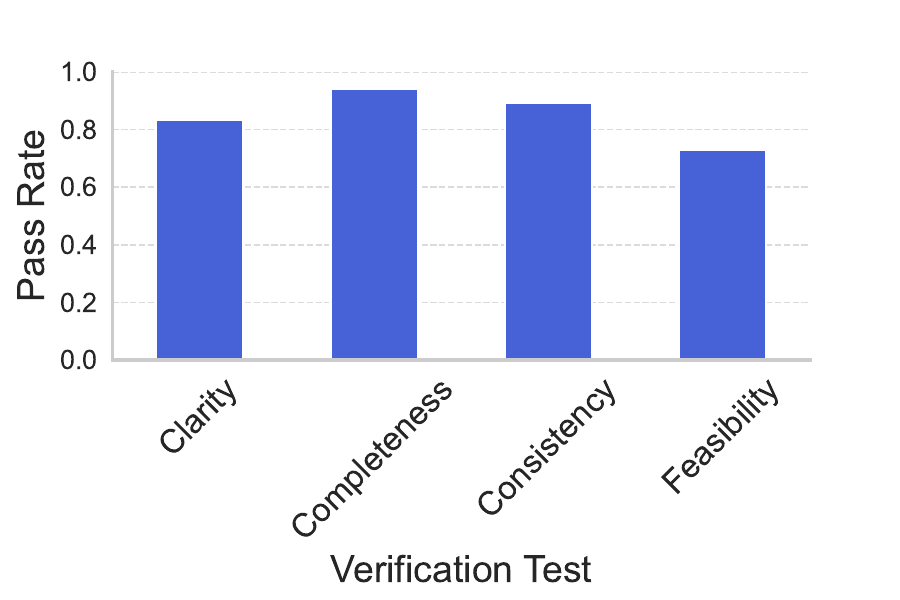}
         \caption{\tasktwo{}}
        \label{fig:verifier_text}
    \end{subfigure}
    \caption{Pass rate for verification checks.}
    \label{fig:verifier}
\end{figure}

Fig.~\ref{fig:verifier_calendar} and Fig.~\ref{fig:verifier_text} show the pass rate for each verification test over all instances generated by \gagent{} for \taskone{} and \tasktwo{} respectively. 


%% file: sections/appendix/_10.5_parameter.tex
\section{\taskone{}'s Parameter Coverage \& Comparison to \textsc{NaturalPlan}}
\label{app:1_coverage}

\begin{figure*}[htbp]
    \centering
    \begin{subfigure}[b]{0.48\textwidth}
        \centering
        \includegraphics[width=\textwidth]{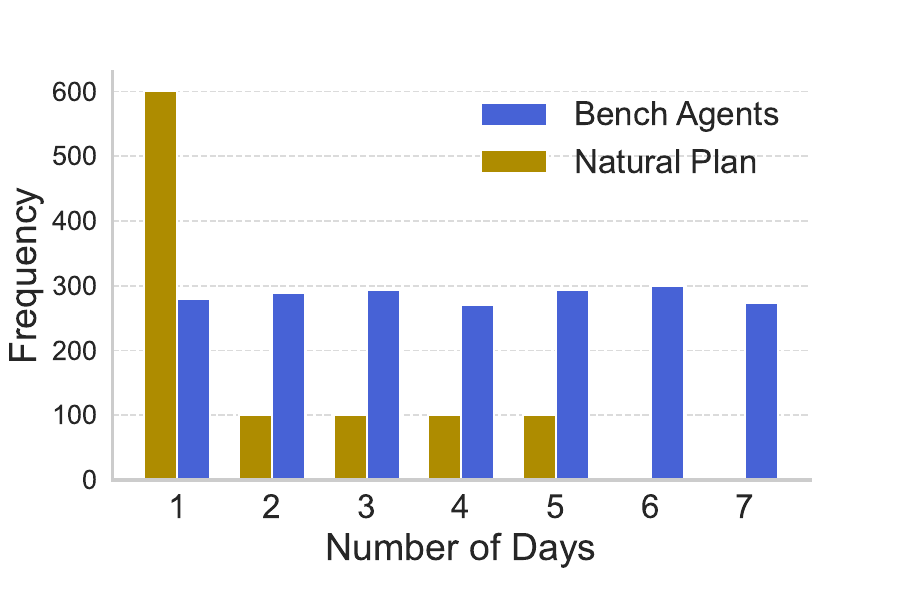}
        \caption{Coverage of Number of Days}
        \label{fig:num_days}
    \end{subfigure}
    \hfill
    \begin{subfigure}[b]{0.48\textwidth}
        \centering
        \includegraphics[width=\textwidth]{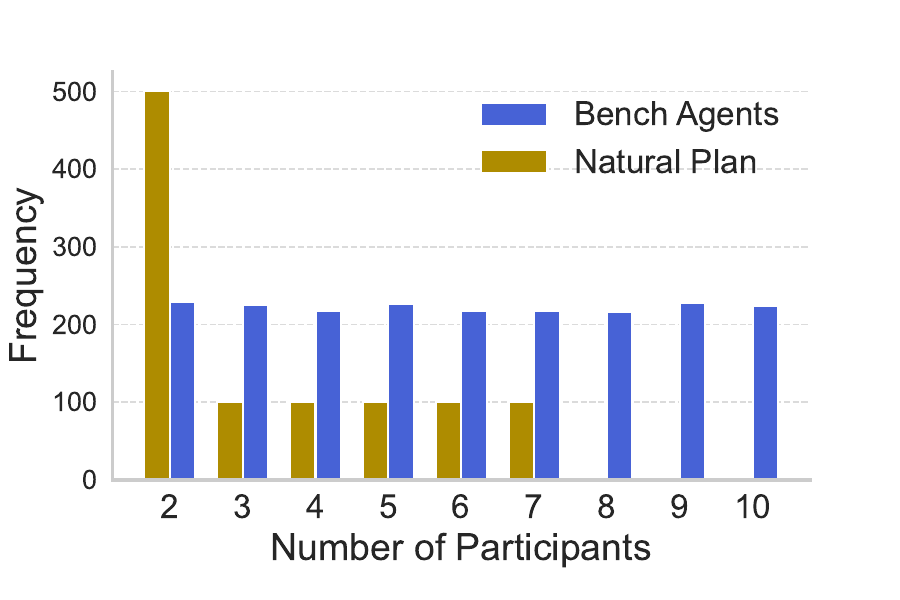}
        \caption{Coverage of Number of Participants }
        \label{fig:num_participants}
    \end{subfigure}
    \caption{Coverage Metrics for Calendar Scheduling Parameters.}
    \label{fig:combined_metrics}
\end{figure*}

\begin{figure}[h!]
    \centering
    \includegraphics[width=0.95\linewidth]{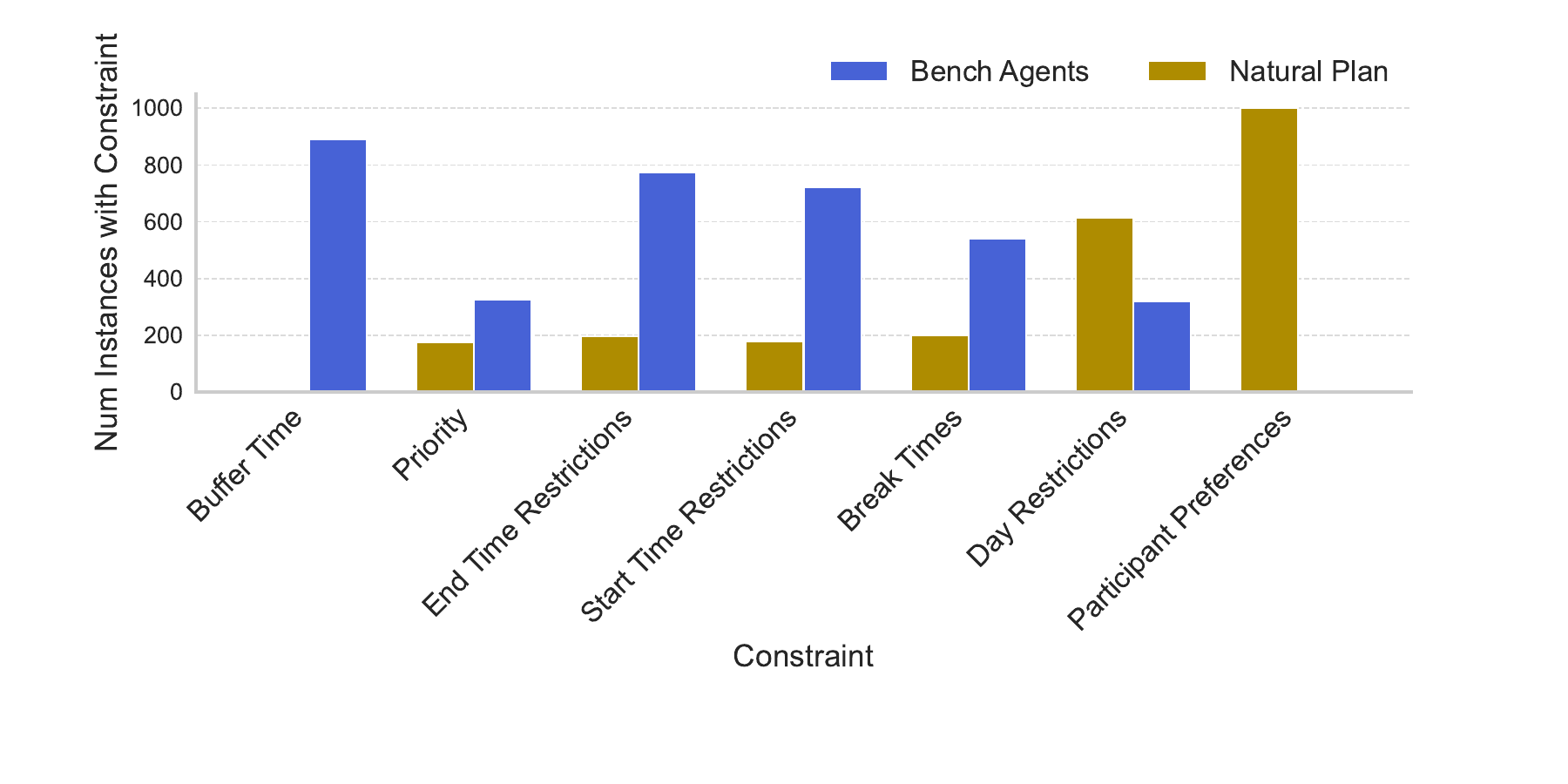}
    \caption{Coverage of Calendar Scheduling Constraints.}
    \label{fig:calendar_constraints}
\end{figure}

For \taskone{}, we may compare to an existing dataset: \textsc{NaturalPlan} \cite{zheng2024naturalplanbenchmarkingllms}. \textsc{NaturalPlan} only includes parameters as metadata, and not constraints. Thus, for the following analysis we extract a set of constraints in each \textsc{NaturalPlan} prompt using GPT-4o. Figures~\ref{fig:num_days} and \ref{fig:num_participants} demonstrate that \sys{} has increased coverage compared to \textsc{NaturalPlan} with respect to the number of days and number of participants parameters. In Figure~\ref{fig:calendar_constraints}, we can see that both \sys{} and \textsc{NaturalPlan} share similar constraints, however, the buffer time constraint is a novel addition by our \pagent{}.

%% file: sections/appendix/_10.6_coverage.tex
\section{\tasktwo{}'s Parameter Coverage}
\label{app:2_coverage}

\begin{figure}[h!]
    \centering
    \includegraphics[width=0.55\linewidth]{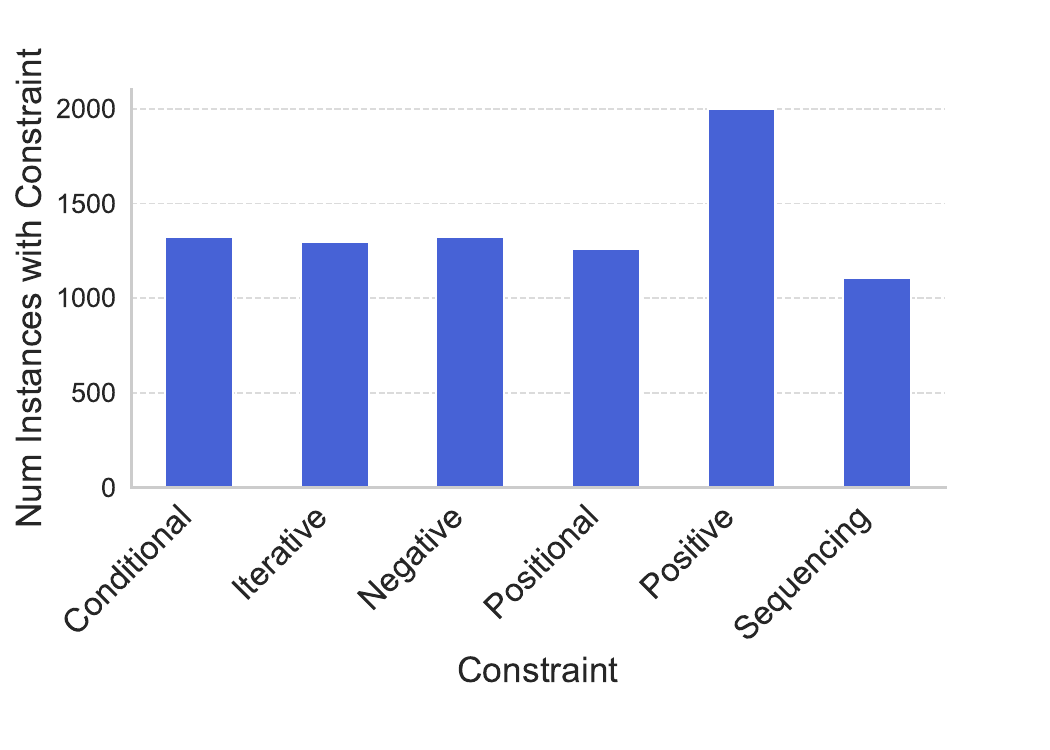}
    \caption{Coverage of Text Generation Constraints.}
    \label{fig:text_constraints}
\end{figure}

In Figure~\ref{fig:text_constraints}, we observe a relatively uniform distribution across constraints. The increased frequency of the positive constraint may be explained by the sampling function of the \gagent{}, whereas, the reduced frequency of positional and sequencing constraints may be explained by the increased likelihood of these two constraints conflicting and so being filtered out by the \vagent{}.

%% file: sections/appendix/_10.3_human.tex
\section{More Details on Human Annotation Experiment}
\label{app:human}

In both the human assessment of the \vagent{} and \eagent{}, annotators were given instructions to annotate each model-based quality check or evaluation metric as either True or False based on the same criteria as output by the \pagent{}. For the human assessment of the \vagent{}, these were clarity, completeness, consistency (and feasibility for \tasktwo{}). For the human assessment of the \eagent{}, definitions of each constraint were given to the annotators in line with the agent configuration and \pagent{} outputs.

\subsection{Human Assessment of \vagent{}}
\label{app:human_verifier}
We conduct a human assessment of the \vagent{} for the model-based checks to evaluate their reliability. In this study, for each generated dataset, we take 50 instances produced from the \gagent{} (before filtering by the \vagent{}). For each instance, we collect two human annotations for each model-based verification check performed (more details in \S~\ref{sec:generation}). 
For the ground truth, we consider an instance to pass a verification check only if both annotators mark it as passed.

\begin{figure}[ht]
    \centering
    \begin{subfigure}[b]{0.4\textwidth}
        \centering
        \includegraphics[width=\textwidth]{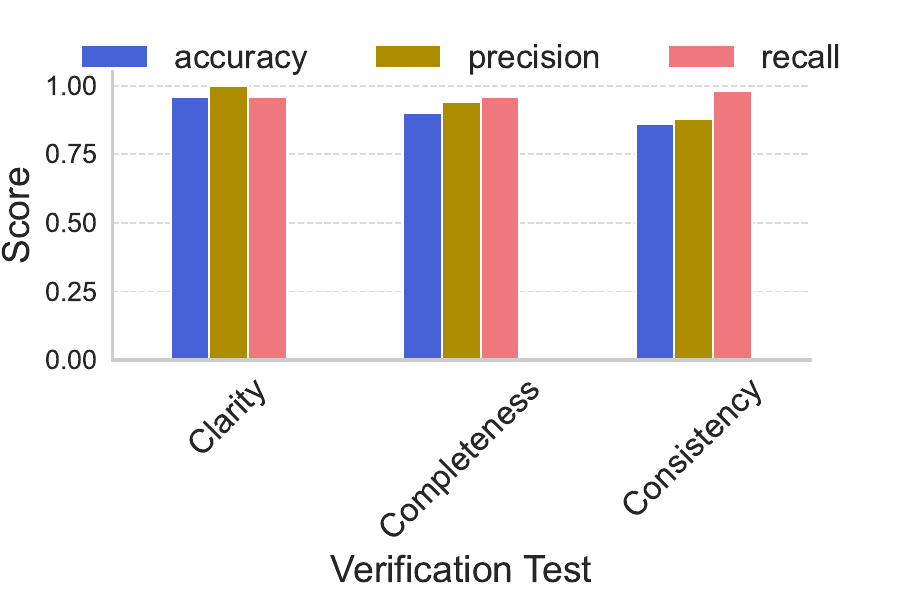}
        \caption{\taskone{}}
        \label{fig:human_eval_verifier_calendar}
    \end{subfigure}
    \hspace{0.05\textwidth} 
    \begin{subfigure}[b]{0.4\textwidth}
        \centering
        \includegraphics[width=\textwidth]{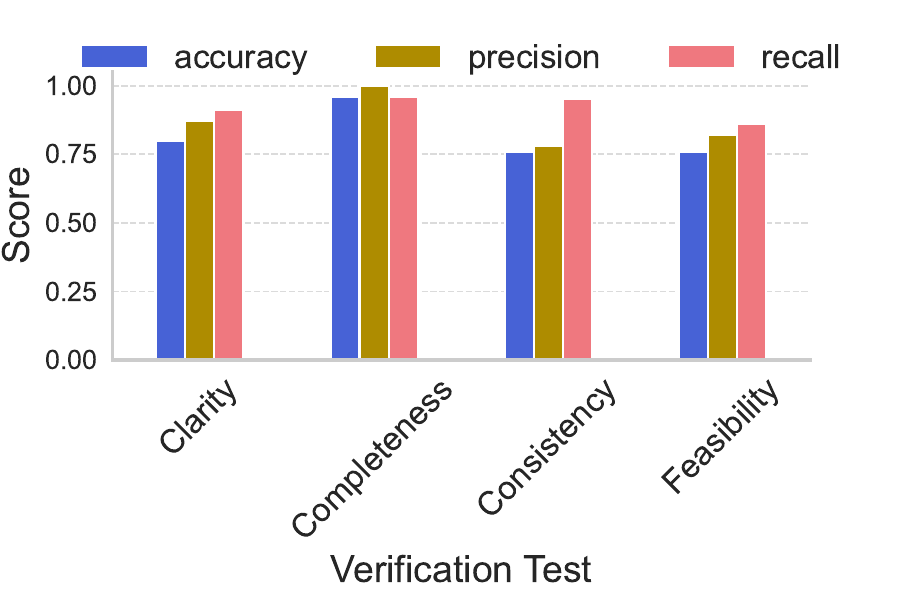}
         \caption{\tasktwo{}}
        \label{fig:human_eval_verifier_text}
    \end{subfigure}
    \caption{\vagent{} scores when comparing them with human annotations.}
    \label{fig:human_eval_verifier}
\end{figure}

Fig.~\ref{fig:human_eval_verifier_calendar} and~\ref{fig:human_eval_verifier_text} report accuracy, precision and recall for \vagent{} and human annotated ground truths.

\subsection{Human Assessment of \eagent{}}
\label{app:eval_check}

\begin{figure}[ht]
\centering
\begin{minipage}{0.3\textwidth}
    \footnotesize
    \centering
    \begin{tabular}{cc}
    \toprule
          &  \tasktwo{} \\ 
    \midrule
    \midrule
    Topic         & 1.00                                       \\
    Positive    & 0.85                                  \\
    Negative     & 0.80                                   \\
    Positional     &   0.90                                 \\
    Sequencing &  0.80                                  \\
    Conditional &  0.75                                  \\
    Iterative &   0.70    \\
    \bottomrule
    \end{tabular}
    \caption{Accuracy for \eagent{} model-based checks.}
    \label{tab:human_evaluator}
\end{minipage}
\hfill
\begin{minipage}{0.65\textwidth}
    \centering
    \includegraphics[width=\textwidth]{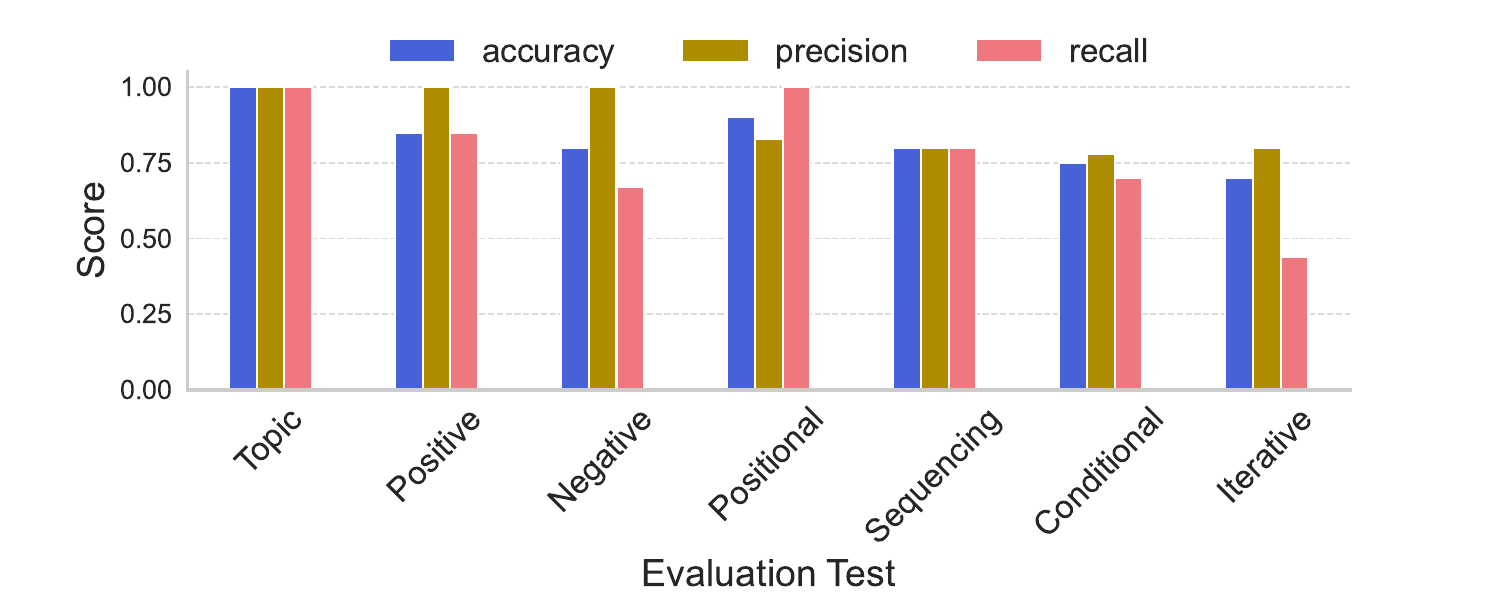}
    \caption{Human annotator and \eagent{} scores for \tasktwo{}.}
    \label{fig:human_eval_text_eval}
\end{minipage}
\end{figure}

We perform a human assessment of the quality of the model-based evaluations for \tasktwo{}. In this study, we take 20 task instances from the \gagent{} after filtering by the \vagent{}, leaving only high quality task instances. For each task instance, we acquire two human annotations for each evaluation test. For the ground truth, we take the worst of both annotations again and our predicted values are obtained by executing the model-based evaluation tests from the \eagent{} on GPT-4o solutions. We report accuracy in Table~\ref{tab:human_evaluator} and precision/recall are reported in Fig.~\ref{fig:human_eval_text_eval}. Finally, we note that the use of an evaluation agent for generative tasks is consistent with prior work \cite{arabzadeh2024assessingverifyingtaskutility}.

%% file: sections/appendix/_10.7_settings.tex
\section{Experimental Settings}
\label{app:setup}

To evaluate each test instance, we perform zero-shot inference with the task prompt, with temperature 0, \texttt{top\_p} 0.95 and \texttt{max\_tokens} 2000. The purported solution is then evaluated under the evaluation criteria provided by the \eagent{}. For each test instance, we obtain a set of evaluation test results for constraint satisfaction.

\subsection{\taskthree{} Text-Only}
\label{sec:text-only}
We experiment with a text-only version of \taskthree{} in which each image is described in text using the object shape, colour, size and x,y co-ordinates from the \texttt{PyMunk} simulation; see Example~\ref{mybox:text_only}.

\begin{mybox}[\taskthree{} Text Only Task Instance]
\label{mybox:text_only}
\textbf{Instructions:} Examine the sequence of 4 images showing the trajectory of objects in a 2D space over time steps $t_1$, $t_2$, $t_3$, $t_4$. Think step by step and describe the changes observed between each consecutive image. Using this information, determine the most likely next state of the system ($t_5$) from the options (a, b, c, d).

\vspace{1em}
\textbf{Image Sequence:}
\begin{itemize}
  \item \textbf{Frame 1}
    \begin{itemize}
      \item Obj 1: purple rectangle with length 76 and height 31 at position (330, 626)
      \item Obj 2: purple square with length 57 at position (470, 111)
      \item Obj 3: green rectangle with length 86 and height 26 at position (172, 383)
      \item Obj 4: red circle with radius 94 at position (595, 391)
      \item Obj 5: orange equilateral triangle with side length 78 at position (507, 627)
      \item Obj 6: red equilateral triangle with side length 38 at position (644, 260)
      \item Obj 7: red rectangle with length 34 and height 15 at position (349, 404)
      \item Obj 8: red equilateral triangle with side length 95 at position (127, 645)
      \item Obj 9: orange square with length 65 at position (247, 412)
    \end{itemize}
  \item \textbf{Frame 2}
    \begin{itemize}
      \item Obj 1: purple rectangle with length 76 and height 31 at position (354, 600)
      \item ...
    \end{itemize}
  \item \textbf{Frame 3}
    \begin{itemize}
      \item Obj 1: purple rectangle with length 76 and height 31 at position (377, 574)
      \item ...
    \end{itemize}
  \item \textbf{Frame 4}
    \begin{itemize}
      \item Obj 1: purple rectangle with length 76 and height 31 at position (399, 547)
      \item ...
    \end{itemize}
\end{itemize}

\vspace{1em}
\textbf{Answer Options (Next State $t_5$):}
\begin{itemize}
  \item \textbf{Option A}
    \begin{itemize}
      \item Obj 1: purple rectangle with length 76 and height 31 at position (500, 175)
      \item ...
    \end{itemize}
  \item \textbf{Option B}
    \begin{itemize}
      \item Obj 1: purple rectangle with length 76 and height 31 at position (420, 519)
      \item ...
    \end{itemize}
  \item \textbf{Option C}
    \begin{itemize}
      \item Obj 1: purple rectangle with length 76 and height 31 at position (458, 460)
      \item ...
    \end{itemize}
  \item \textbf{Option D}
    \begin{itemize}
      \item Obj 1: purple rectangle with length 76 and height 31 at position (536, 300)
      \item ...
    \end{itemize}
\end{itemize}

\vspace{0.5em}
\textbf{Provide your response as:} \texttt{Changes: <describe the changes> Answer: <letter>}
\end{mybox}

%% file: sections/appendix/_10.8_additional.tex
\newpage
\section{Additional Results on Model Analysis (\S~\ref{sec:final_results})}
\label{sec:additional_results}

\begin{figure}[h!]
    \centering
    
    \begin{subfigure}[b]{0.45\textwidth}
        \centering
        \includegraphics[width=\textwidth]{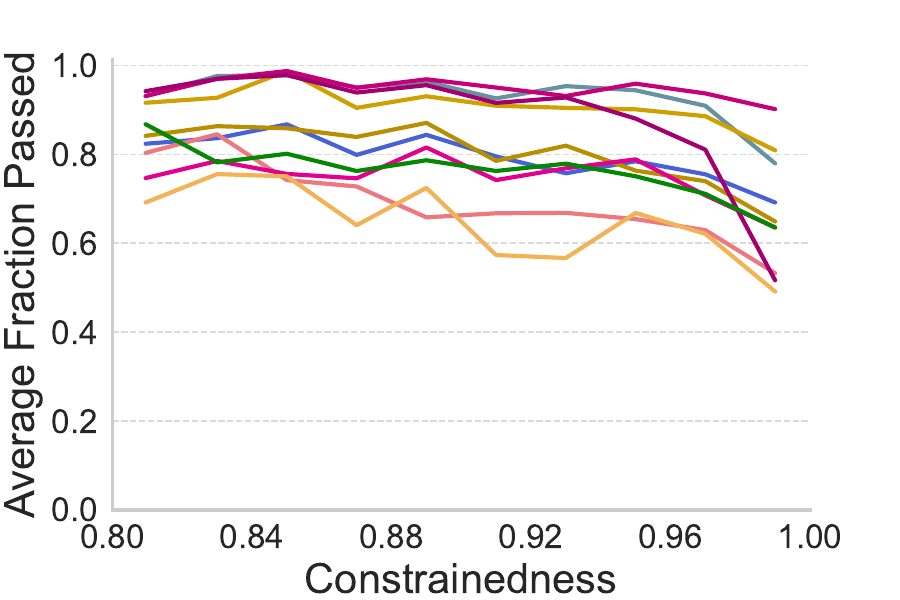}
        \caption{\taskone{} (Fraction Passed)}
        \label{fig:calendar_fraction_constrainedness}
    \end{subfigure}
    \vspace{0.5cm}  
    \begin{subfigure}[b]{0.45\textwidth}
        \centering
        \includegraphics[width=\textwidth]{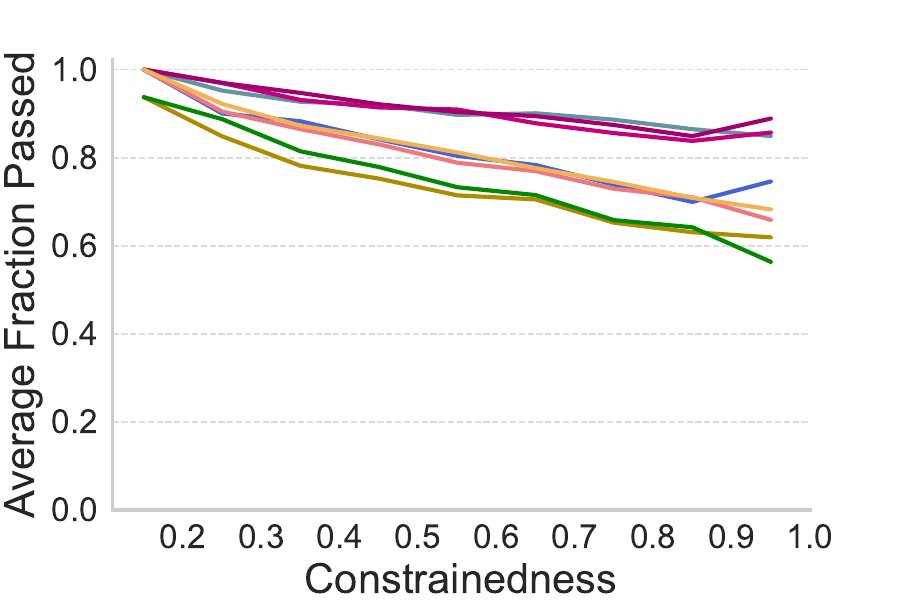}
        \caption{\tasktwo{} (Fraction Passed)}
        \label{fig:text_fraction_constrainedness}
    \end{subfigure}
    \begin{tabular}{c}  
        \includegraphics[width=0.95\textwidth]{figures/legends/legend_image_calendar_rows2_calendar_rows2.pdf} \\
    \end{tabular}
    \caption{Average Fraction Passed with Increasing Constrainedness.}
    \label{fig:fraction_constrainedness_plots}
\end{figure}

Here, we present some additional results from our model analysis in \S~\ref{sec:final_results}. Fig.\ref{fig:fraction_constrainedness_plots} reports the average fraction passed with increasing constrainedness for \taskone{} and \tasktwo{}. Further, Fig.\ref{fig:params_by_pass_all} shows the average pass all with increasing parameter ranges for the number of days and number of participants parameters in \taskone{}.

\input{figures/tex/combined_evaluator_not_pass_all}

%% file: figures/tex/combined_evaluator_not_pass_all.tex
\begin{figure*}[ht]
    \centering
    \begin{subfigure}[b]{0.45\textwidth}
        \centering
        \includegraphics[width=\textwidth]{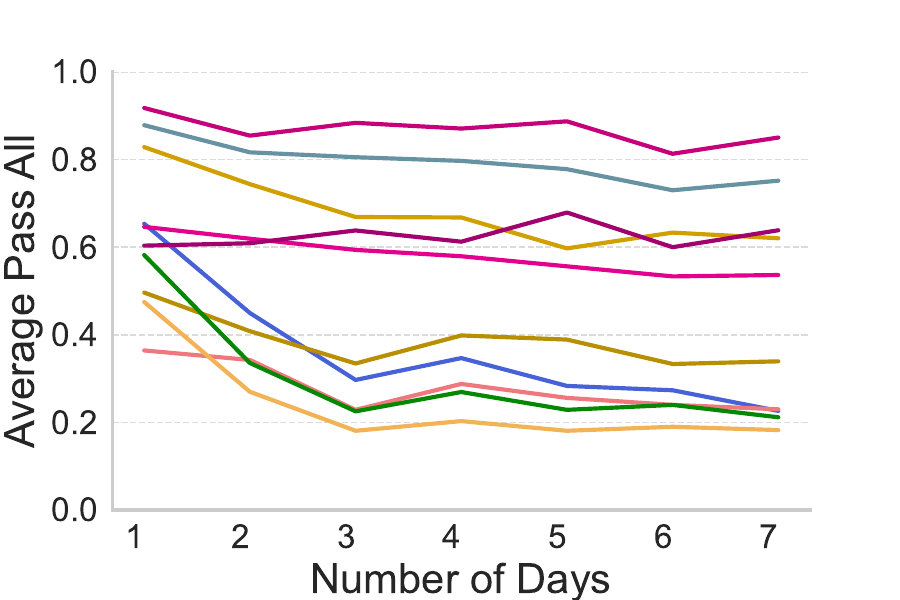}
        \caption{\taskone{}} 
        \label{fig:pass_rate_num_days}
    \end{subfigure}
    \begin{subfigure}[b]{0.45\textwidth}
        \centering
        \includegraphics[width=\textwidth]{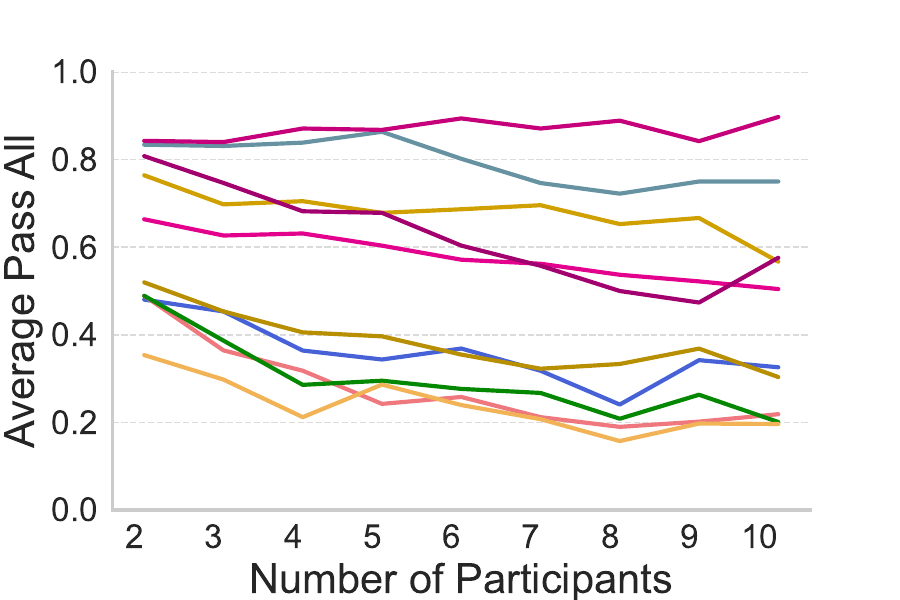}
        \caption{\taskone{}}
        \label{fig:pass_rate_num_participants}
    \end{subfigure}
    \begin{tabular}{c}  
        \includegraphics[width=\textwidth]{figures/legends/legend_image_calendar_rows2_calendar_rows2.pdf} \\
    \end{tabular}

    \caption{Model performance with increasing availability schedule complexity.}
    \label{fig:params_by_pass_all}
\end{figure*}

%% file: sections/appendix/_10.9_quantifying.tex
\newpage
\section{Quantifying Developer Feedback}
\label{app:feedback}

\input{figures/tex/distance}

From \S~\ref{sec:details}, recall that developer feedback can utilize the outputs of different agents to refine generation. To quantify the amount of developer-in-the-loop feedback, we calculate the Levenshtein distance on the natural language plans generated by  \pagent{} and the code generated by the respective \gagent{}, \vagent{}, and \eagent{}. In each case, we calculate the distance between the agent-generated plan or code and the plan or code {\em after} developer-in-the-loop edits and normalize with the maximum number of characters. 

\noindent{\bf Observations:} Table~\ref{tab: edit_distance} reports the normalized distance and the maximum number of characters for each dataset. We observe that \taskone{} requires substantially more characters overall and proportionally more developer-in-the-loop feedback compared to \tasktwo{}. We pose that this is because more intervention is needed to ensure the correct parsing of logical conditions verified and evaluated using programmatic tests; constraints in \tasktwo{} are not as challenging to parse. This in turn results in better data generated for \taskone{}(\S~\ref{sec:quality}). BA-Causal presents a distinct pattern: while the plan required relatively less developer edits (0.00), reflecting the well-defined nature of physics simulation parameters, substantial intervention was needed for G-Agent code generation (0.27) to correctly implement the PyMunk simulation and trajectory sampling logic. The V-Agent plan also required significant refinement (0.34) to properly define feasibility checks for physics-based instances, such as ensuring sufficient separation between answer and distractor frames. This pattern illustrates that for tasks leveraging external libraries and simulations, the primary developer effort shifts from conceptual planning to ensuring correct implementation and defining appropriate quality criteria for deterministically generated data.

%% file: figures/tex/distance.tex
\begin{table*}[h!]
    \centering
    \small
\begin{tabular}{lcccccc}
    \toprule
    \multicolumn{1}{c}{\bf Dataset} & \multicolumn{2}{c}{\gagent{}} & \multicolumn{2}{c}{\vagent{}} & \multicolumn{2}{c}{\eagent{}}\\
    & Plan & Code & Plan & Code & Plan & Code\\
    \midrule
    \midrule
    \taskone{} & 0.34 (827) & 0.14 (11,491) & 0.41 (911) & 0.34 (11,399) & 0.0 (608) & 0.25 (17,449) \\
    \tasktwo{} & 0.02 (3,093) & 0.15 (6,145) & 0.15 (687) & 0.05 (6,276) & 0.25 (700) & 0.08 (7,687) \\
    \taskthree{} & 0.00 (3,061) & 0.27 (10,472) & 0.34 (712) & 0.19 (5,721) & -- & -- \\
    \bottomrule
\end{tabular}
    \caption{Normalized Levenshtein distance (max number of characters) between model generations pre- and post- developer feedback.}
    \label{tab: edit_distance}
\end{table*}

%% file: sections/appendix/_10.11_full_eval.tex
\section{Full Evaluation Plots}
\label{app:full_plots}

\input{figures/tex/ba_calendar_master}

\input{figures/tex/ba_text_master}

\input{figures/tex/ba_causal_master}

%% file: figures/tex/ba_calendar_master.tex
\begin{figure*}[ht]
    \centering
    \begin{tabular}{ccc}  
        \begin{subfigure}[b]{0.32\textwidth}
            \centering
            \includegraphics[width=\textwidth, height=3.5cm, keepaspectratio]{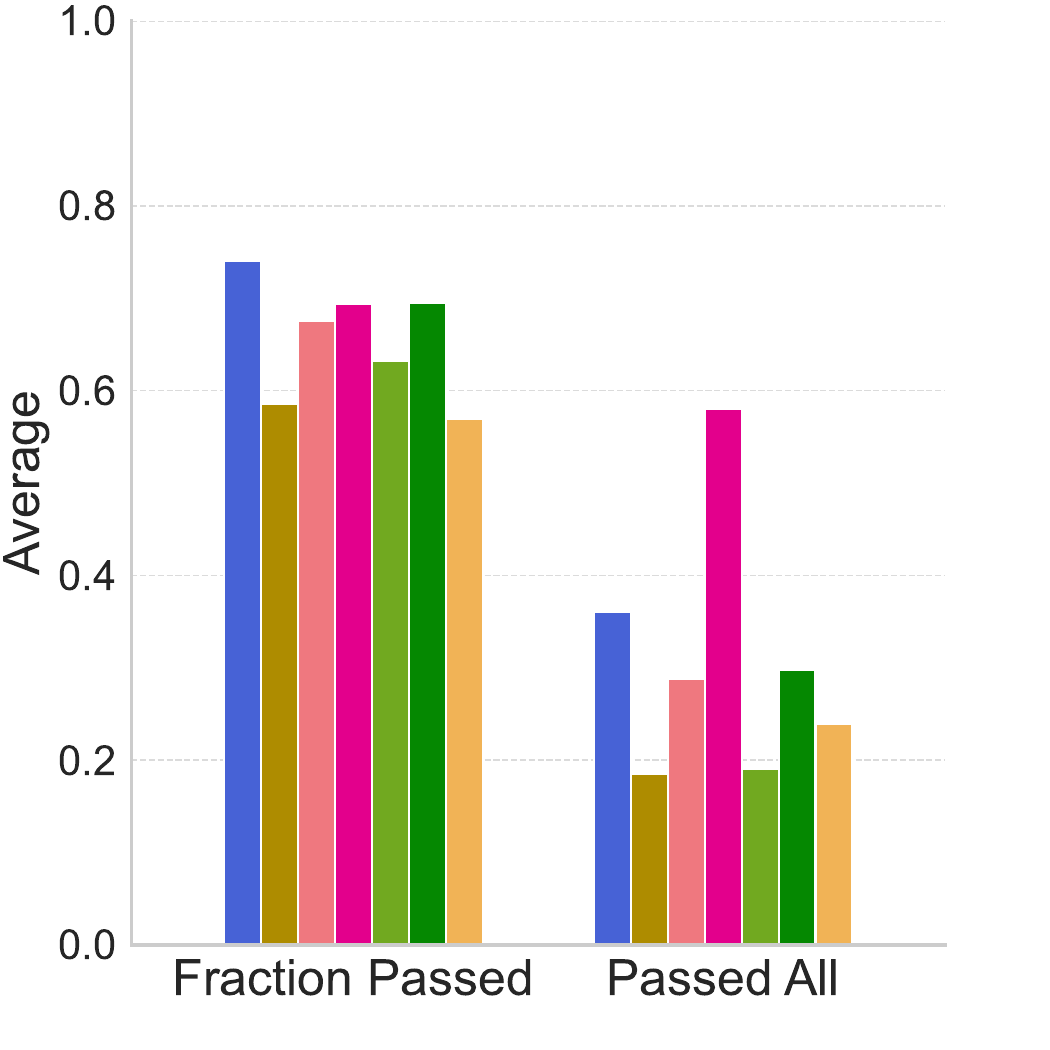}
            \captionsetup{font={scriptsize}}
            \caption{Performance}
            \label{fig:master_calendar_results}
        \end{subfigure}
        &
        \begin{subfigure}[b]{0.32\textwidth}
            \centering
            \includegraphics[width=\textwidth, height=3.5cm, keepaspectratio]{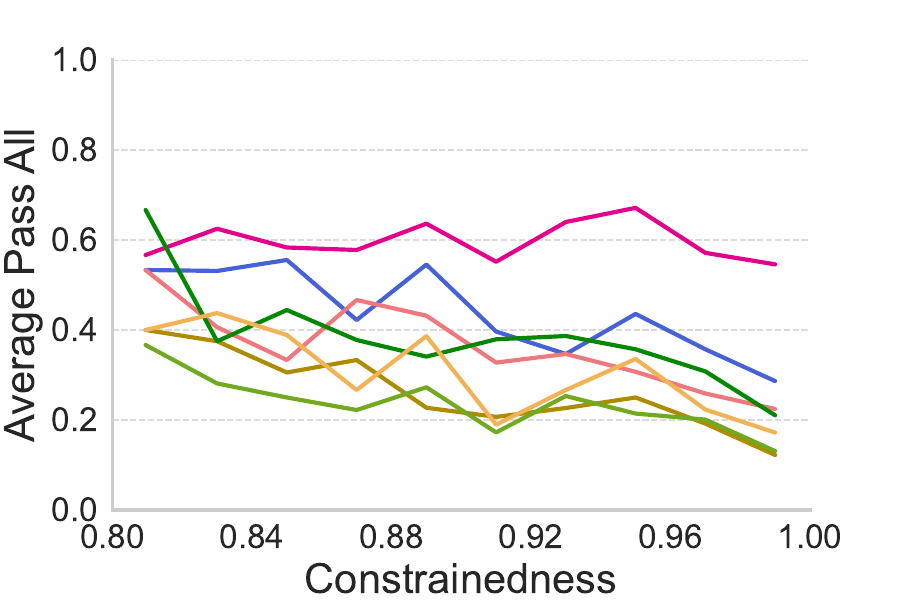}
            \captionsetup{font={scriptsize}}
            \caption{Accuracy v/s Constrainedness}
            \label{fig:master_calendar_accuracy_vs_constrainedness}
        \end{subfigure}
        &
        \begin{subfigure}[b]{0.32\linewidth}
            \centering
            \includegraphics[width=\linewidth, height=3.5cm, keepaspectratio]{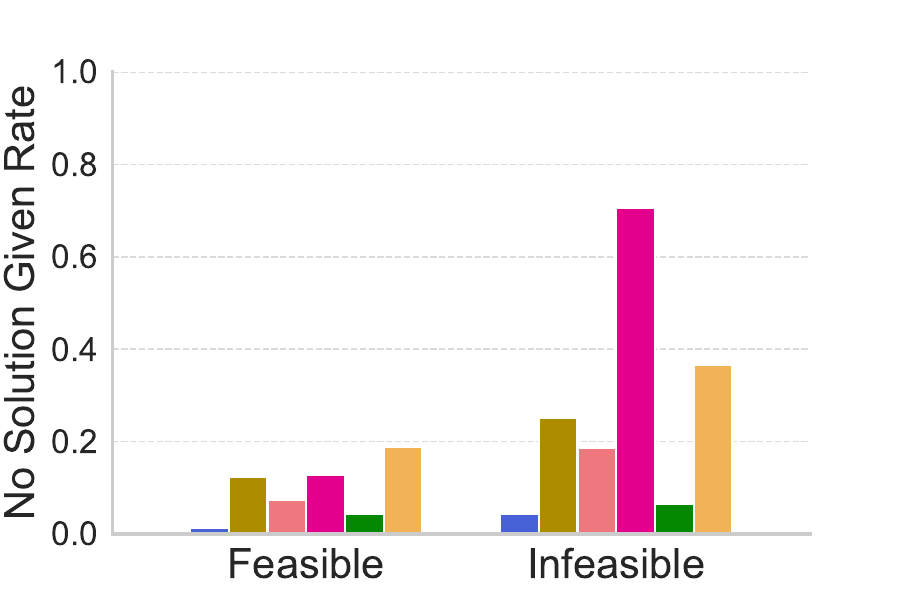}
            \captionsetup{font={scriptsize}}
            \caption{Refusal Rate}
            \label{fig:master_calendar_refusal}
        \end{subfigure}
        \vspace{-4.5pt}
    \\
        \multicolumn{3}{c}{
            \begin{tabular}{cc}
                \begin{subfigure}[b]{0.33\textwidth}
                    \centering
                    \includegraphics[width=\textwidth, height=3.5cm, keepaspectratio]{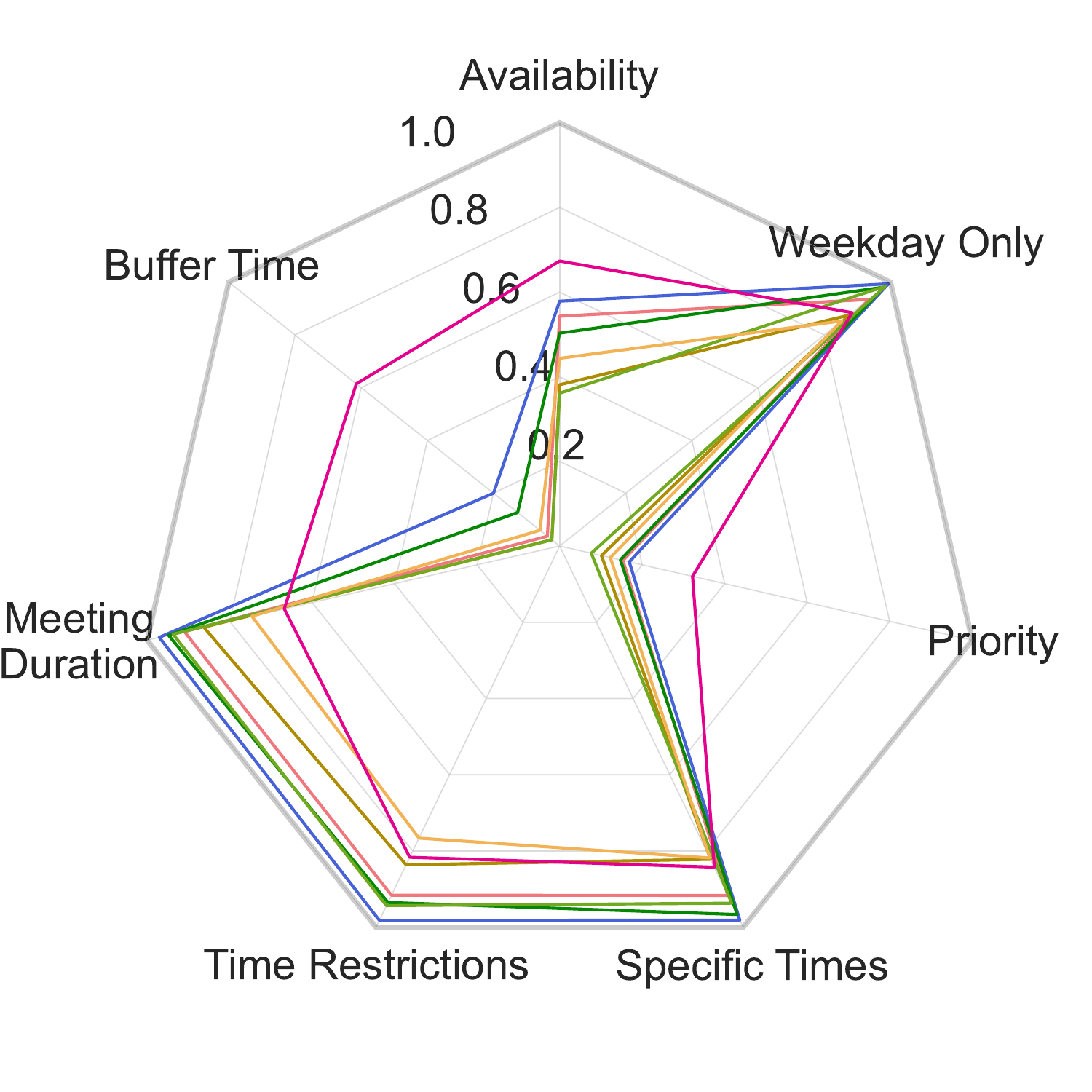}
                    \captionsetup{font={scriptsize}}
                    \caption{By Constraint}
                    \label{fig:master_calendar_results_by_constraint}
                \end{subfigure}
                &
                \begin{subfigure}[b]{0.33\textwidth}
                    \centering
                    \includegraphics[width=\textwidth, height=3.5cm, keepaspectratio]{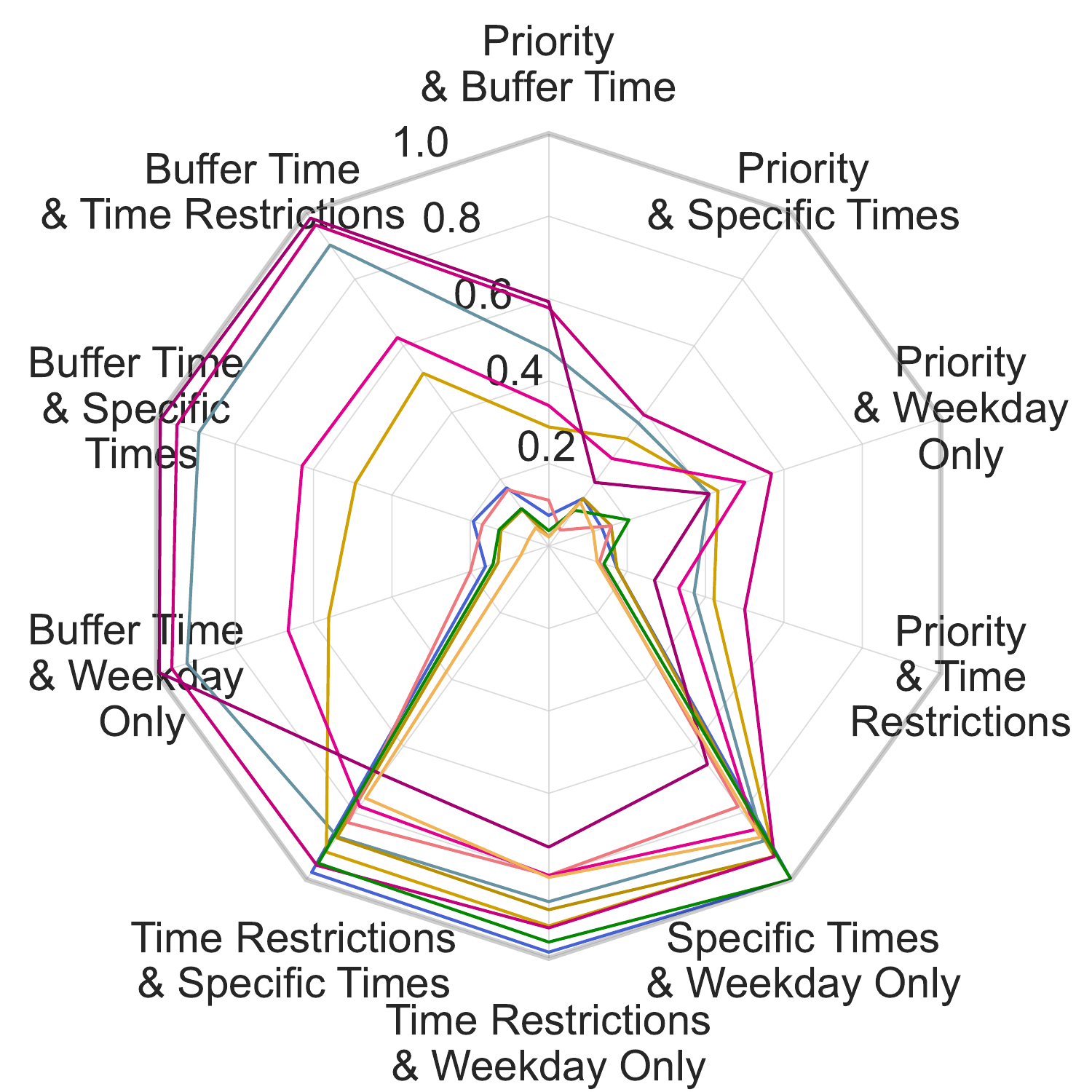}
                    \captionsetup{font={scriptsize}}
                    \caption{By Pairwise}
                    \label{fig:master_calendar_pairwise_pass_rate}
                \end{subfigure}
            \end{tabular}
        }
    \end{tabular}
    
    \centering
    \begin{tabular}{c}  
    \includegraphics[width=0.92\textwidth]{figures/legends/legend_image_calendar_rows2_calendar_rows2.pdf} \\
    \end{tabular}
    \caption{\footnotesize \taskone{}: (a) fraction passed and pass all for all task instances; (b) pass all accuracy vs. constrainedness (c) rate of `no solution' outcomes (lower is better for feasible, higher for infeasible); (d, e) pass rate for a given constraint(s) for task instances where constraints are applied}
    \label{fig:master_combined_evaluator_plots}
    \vspace{0mm}
\end{figure*}

%% file: figures/tex/ba_text_master.tex
\begin{figure*}[t]
    \centering
    \begin{tabular}{ccc}  
        \begin{subfigure}[b]{0.27\textwidth}
            \centering
            \includegraphics[width=\textwidth]{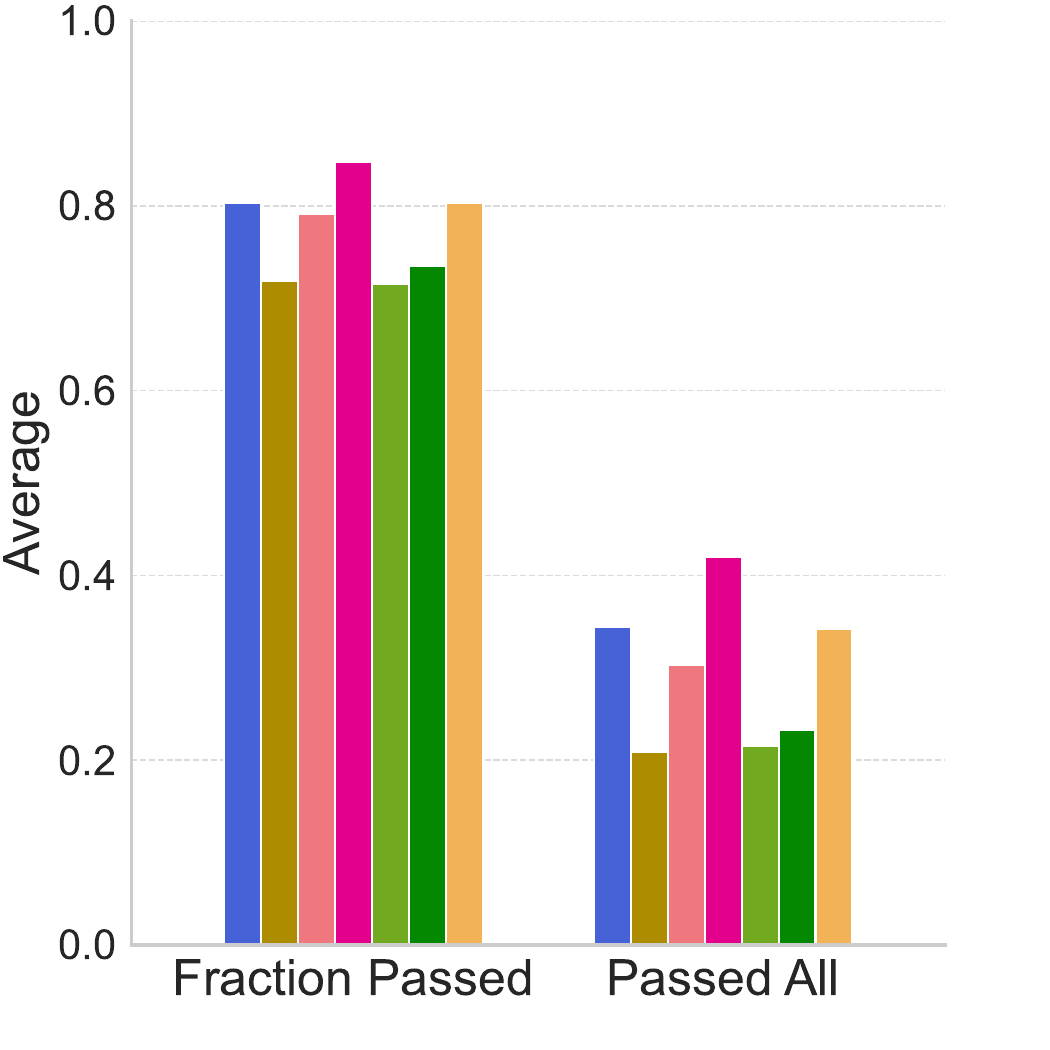}
            \captionsetup{}
            \caption{Performance}
            \label{fig:master_text_results}
        \end{subfigure}
        &
        \begin{subfigure}[b]{0.27\textwidth}
            \centering
            \includegraphics[width=\textwidth]{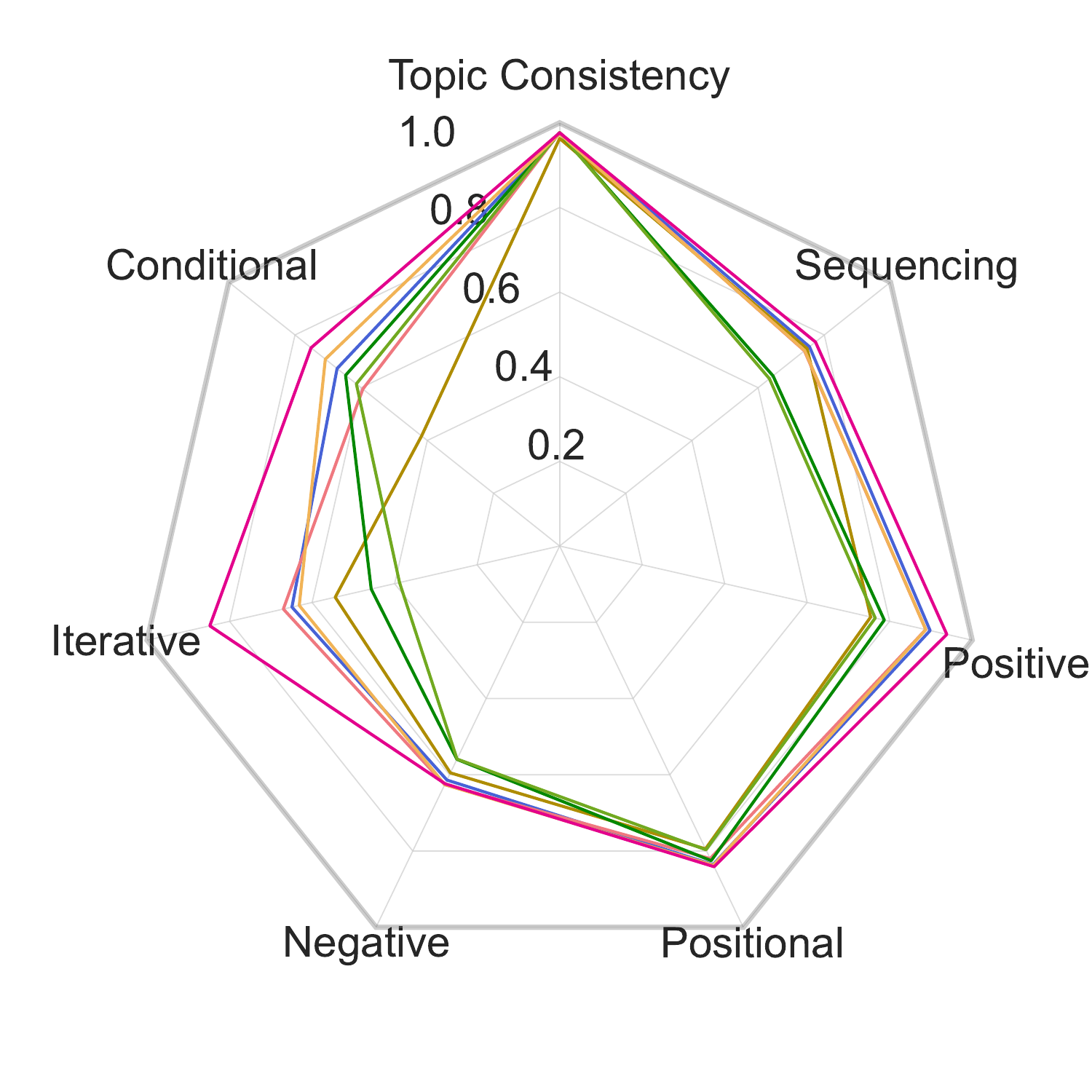}
            \captionsetup{}
            \caption{By Constraint}
            \label{fig:master_text_results_by_constraint}
        \end{subfigure}
        &
        \begin{subfigure}[b]{0.24\textwidth}
            \centering
            \includegraphics[width=\textwidth]{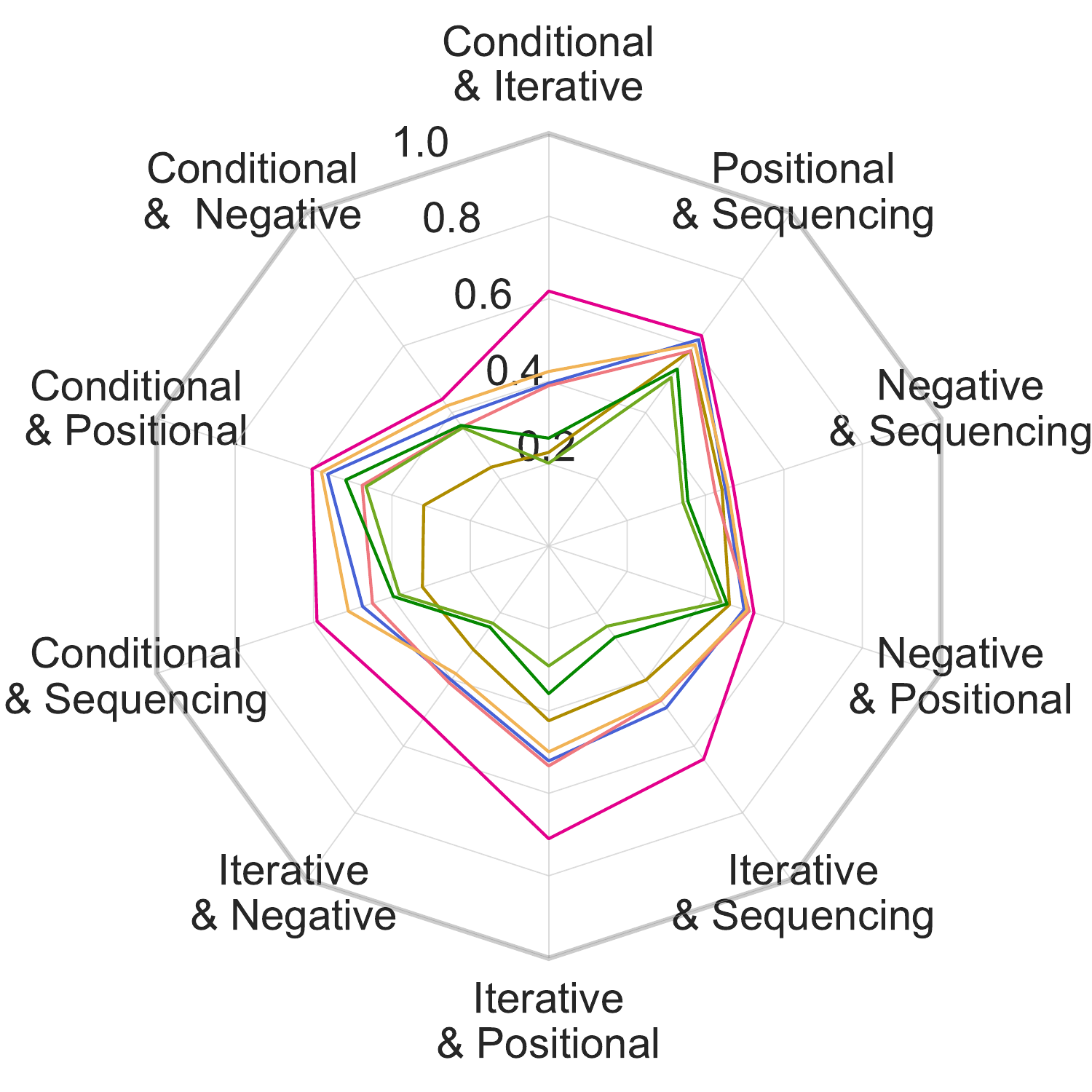}
                    \vspace{-2pt}
            \captionsetup{}
            \caption{By Pairwise}
            \label{fig:master_text_pairwise_pass_rate}
        \end{subfigure}
        \vspace{-4.5pt}
    \end{tabular}
    
    \centering
    \begin{tabular}{c}  
        \includegraphics[width=0.72\textwidth]{figures/legends/legend_image_text_rows2_text_rows2.pdf} \\
    \end{tabular}
    \caption{\footnotesize \tasktwo{}: (a) show fraction passed and pass all for all task instances; (c) and (d) show pass rate for a given constraint(s) for task instances where constraints are applied\vspace{-8mm}}
    \label{fig:master_combined_evaluator_plots}
    
\end{figure*}

%% file: figures/tex/ba_causal_master.tex
\begin{figure*}[ht]
    \centering
    \begin{tabular}{ccc}  
        \begin{subfigure}[b]{0.3\textwidth}
            \centering
            \includegraphics[width=\linewidth]{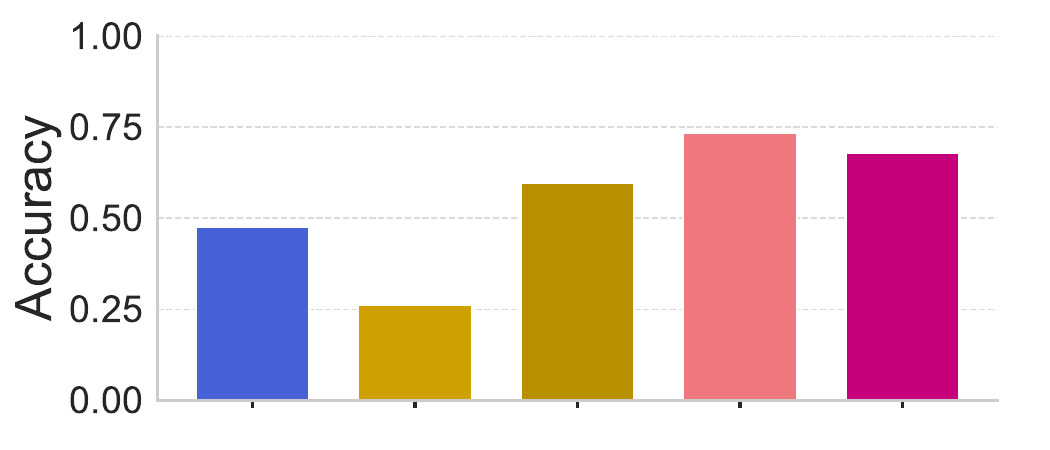}
        \captionsetup{}
        \caption{ Performance}    \label{fig:master_causal_results}
        \end{subfigure}
        &
        \begin{subfigure}[b]{0.3\textwidth}
            \centering
            \includegraphics[width=\textwidth]{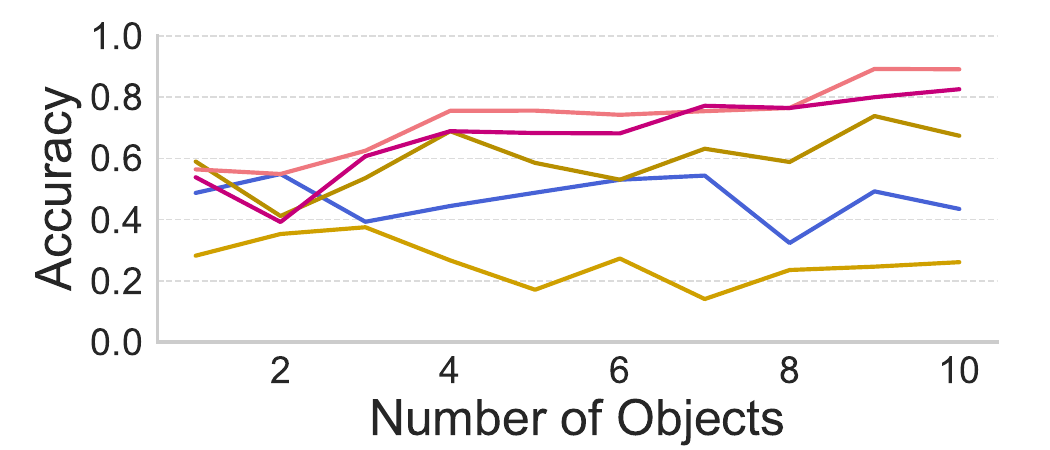}
        \captionsetup{}
        \caption{\# of Objects }
            \label{fig:master_causal_accuracy_vs_objects}
        \end{subfigure}
        &
    \begin{subfigure}{0.3\linewidth}
        \centering
        \includegraphics[width=\linewidth]{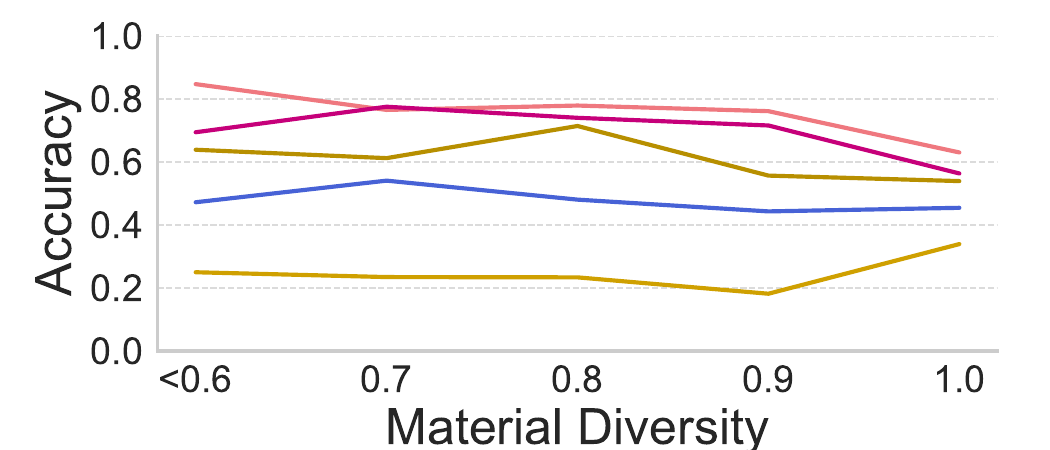}
        \caption{Material Diversity}
        \label{fig:master_causal_accuracy_vs_material}
    \end{subfigure}
    
        \vspace{-4.5pt}
    \\
    \begin{subfigure}[b]{0.3\textwidth}
            \centering
            \includegraphics[width=\linewidth]{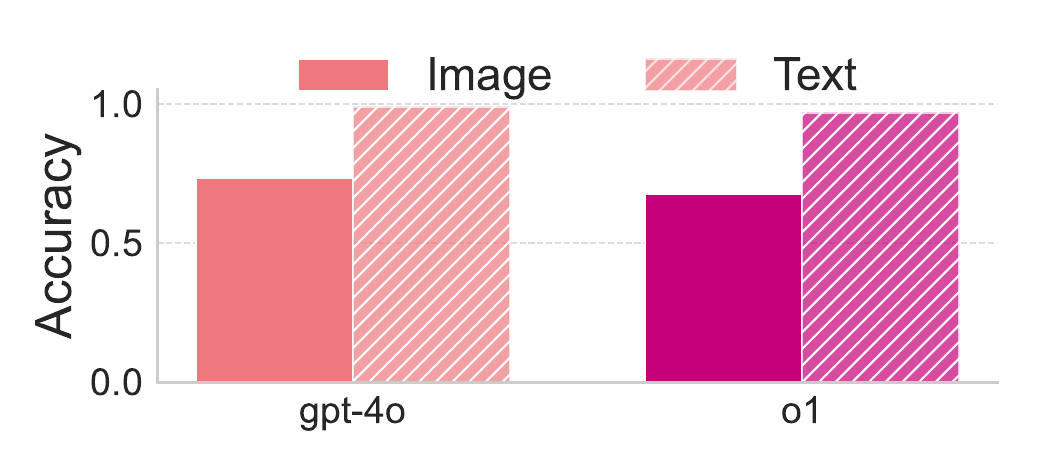}
            \captionsetup{}
            \caption{Image v/s Text Input}
            \label{fig:master_causal_text_image}
        \end{subfigure}
        &
        
    \begin{subfigure}{0.3\linewidth}
        \centering
        \includegraphics[width=\linewidth]{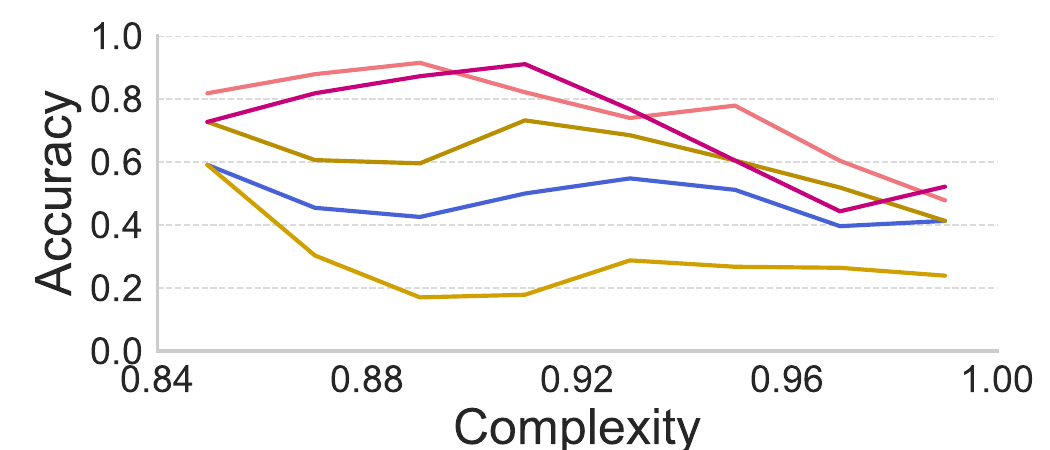}
        \caption{Accuracy v/s Complexity}
        \label{fig:master_causal_complexity_results}
    \end{subfigure}
    &
    \begin{subfigure}{0.3\linewidth}
        \centering
        \includegraphics[width=\linewidth]{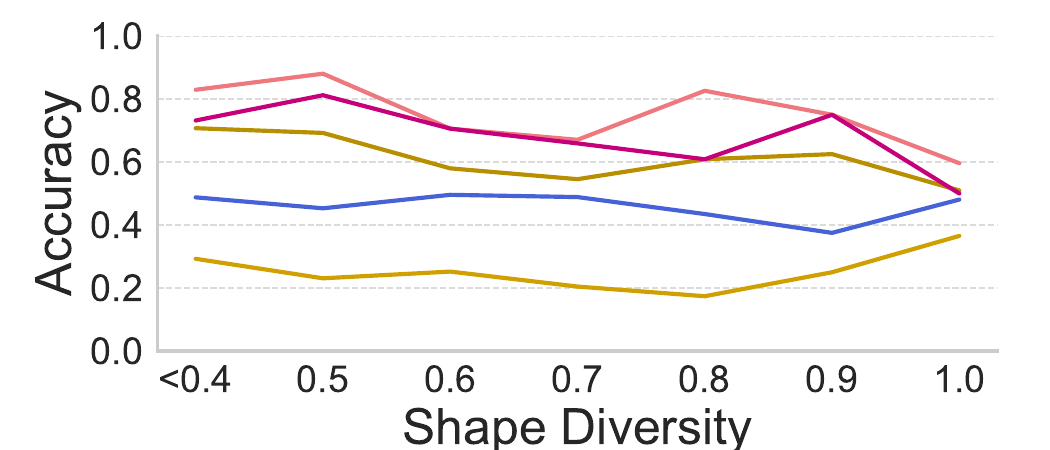}
        \caption{Shape Diversity}
        \label{fig:master_causal_accuracy_vs_shape}
    \end{subfigure}
    \end{tabular}
    \centering
    \begin{tabular}{c}  
        \includegraphics[width=0.85\textwidth]{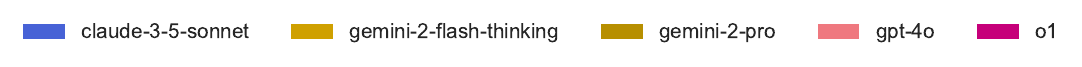} \\
    \end{tabular}
    \caption{\footnotesize \taskthree{}: (a) overall accuracy; accuracy by (b) number of objects, (c) material diversity, (e) complexity, and (f) shape diversity; (d) compares text-only vs. original image task.}
\label{fig:master_combined_evaluator_plots}
    \vspace{-15pt}
\end{figure*}